\documentclass[]{interact}

\usepackage{epstopdf}
\usepackage{subfigure}

\usepackage{natbib}
\bibpunct[, ]{(}{)}{;}{a}{}{,}
\renewcommand\bibfont{\fontsize{10}{12}\selectfont}

\theoremstyle{plain}

\theoremstyle{definition}

\theoremstyle{remark}

\usepackage{todonotes}

\usepackage{siunitx}
\usepackage{soul,framed}
\usepackage{color}
\usepackage[normalem]{ulem} 

\usepackage{colortbl}
\usepackage{arydshln}
\usepackage[flushleft]{threeparttable}
\usepackage{algorithm,algorithmic}
\usepackage{pifont}
\newcommand{\cmark}{\ding{51}}%
\newcommand{\xmark}{\ding{55}}%

\usepackage{makecell}

\definecolor{darkgreen}{rgb}{0,0.6,0.2}

\begin{document}


\title{Multi-modal land cover mapping of remote sensing images using pyramid attention and gated fusion networks}

\author{
\name{Qinghui Liu\textsuperscript{a,b}\thanks{CONTACT Qinghui Liu. Email: samleoqh@gmail.com} , Michael Kampffmeyer\textsuperscript{b,a}, Robert Jenssen\textsuperscript{b,a} and Arnt-B{\o}rre Salberg\textsuperscript{a}}
\affil{\textsuperscript{a}Norwegian Computing Center, Dept. SAMBA, P.O. Box 114 Blindern, NO-0314 Oslo, Norway; \textsuperscript{b}UiT Machine Learning Group, Department of Physics and Technology, UiT the Arctic University of Norway, Troms{\o}, Norway}
\thanks{All the authors are associated with the Centre for Research-based Innovation \emph{Visual Intelligence}: \texttt{http://visual-intelligence.no}, funded by the Research Council of Norway and consortium partners. RJ, MK and QL are with the UiT Machine Learning Group: \texttt{http://machine-learning.uit.no}.}
}

\maketitle

\begin{abstract}
Multi-modality data is becoming readily available in remote sensing (RS) and can provide complementary information about the Earth's surface. Effective fusion of multi-modal information is thus important for various applications in RS, but also very challenging due to large domain differences, noise, and redundancies.
There is a lack of effective and scalable fusion techniques for bridging multiple modality encoders and fully exploiting complementary information. To this end, we propose a new multi-modality network (MultiModNet) for land cover mapping of multi-modal remote sensing data based on a novel pyramid attention fusion (PAF) module and a gated fusion unit (GFU). The PAF module is designed to efficiently obtain rich fine-grained contextual representations from each modality with a built-in cross-level and cross-view attention fusion mechanism, and the GFU module utilizes a novel gating mechanism for early merging of features, thereby diminishing hidden redundancies and noise. This enables supplementary modalities to effectively extract the most valuable and complementary information for late feature fusion. Extensive experiments on two representative RS benchmark datasets demonstrate the effectiveness, robustness, and superiority of the MultiModNet for multi-modal land cover classification.
\end{abstract}

\section{Introduction}
Automatic mapping of land cover using remote sensing (RS) data is of great importance for a wide range of earth observation applications since it provides a fast and cost-effective solution for analyzing large areas ~\citep{ab_00, audebert2016semantic}. This includes applications like urban planning~\citep{noor2018remote}, precision agriculture~\citep{liu_00, Liu_2020_CVPR_Workshops}, and disaster management~\citep{ab_01, bello2014satellite, fan2021disaster}, to name a few. In the past few years, the emergence of deep learning and convolutional neural networks (CNNs) has led to significant improvements for land cover mapping in RS~\citep{maggiori2017can, audebert2018beyond, pashaei2020review, liu2020TGRS}. Many existing deep learning approaches, however, only use unimodal remote sensing images, e.g., the standard three-channel data such as RGB or IRRG (NIR-Red-Green) images. Multi-modality data is becoming readily available and increasingly essential in remote sensing.
This raises open challenges such as "what," "how," and "where" to effectively fuse multi-modal data~\citep{hong2020more} in order to develop joint representations of multiple modalities for enhancing land cover mapping performance.

\begin{figure}[htbp]
 \centering
  \includegraphics[width=0.8\columnwidth]{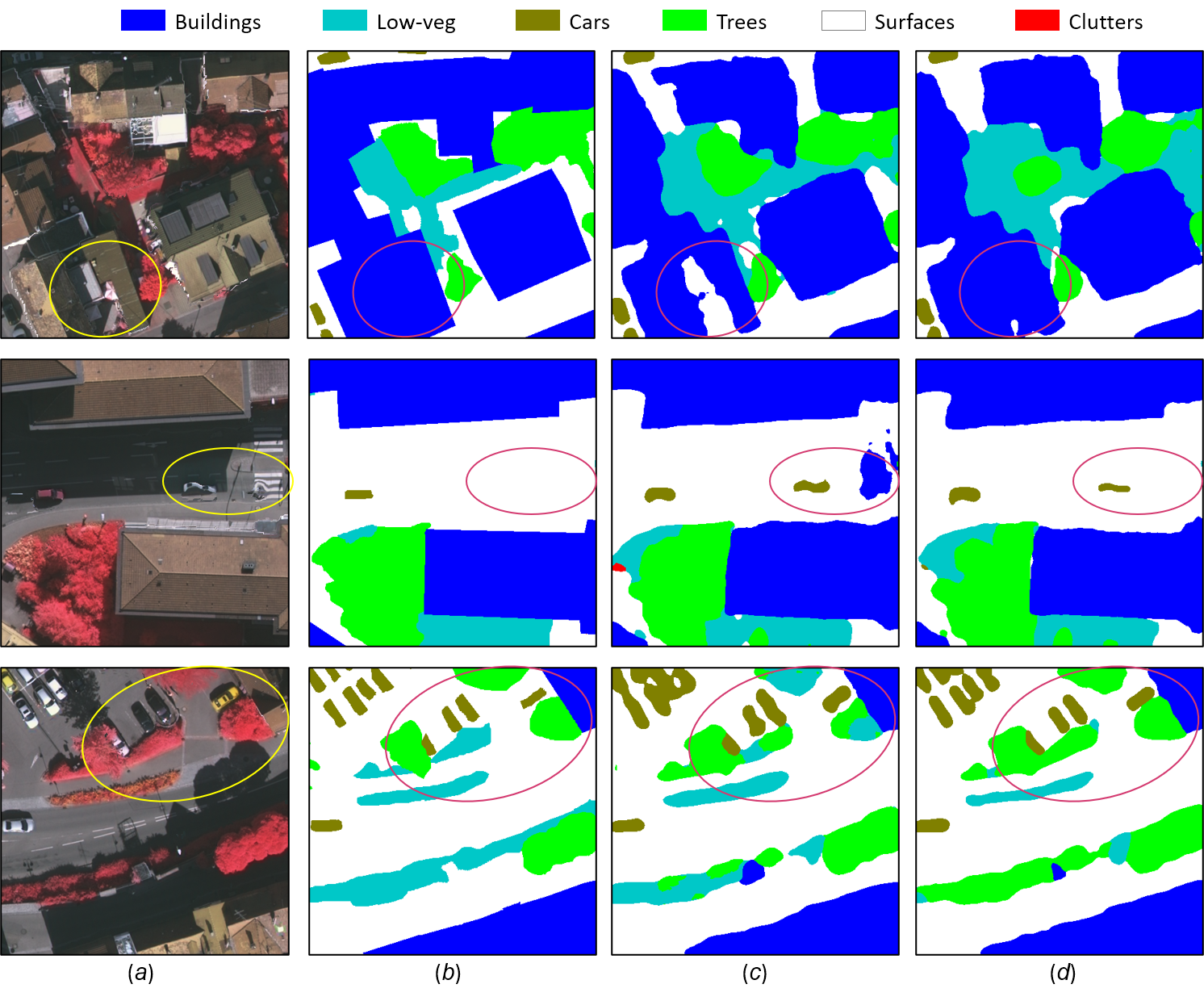} 
  \caption{Mismapped or mislabeled examples in the Vaihingen dataset. (a) the IRRG images, (b) the labels, (c) the mapping results from a unimodal (only IRRG) model, and (d) the mapping results from our multi-modal (IRRG + DSM) model.}
  \label{fig:mismap}
\end{figure}

Remote sensing imagery is often characterized by complex data properties in the form of heterogeneity and class imbalance, as well as overlapping class-conditional distributions that bring severe challenges for generating land cover maps or detecting and localizing objects, producing a high degree of uncertainty in obtained results. As shown in Fig.~\ref{fig:mismap}, mismapped or mislabeled results appear in the unimodal case for objects with similar color and texture, e.g., the roof of buildings vs.\ surfaces, and, the trees vs.\ low vegetation in Vaihingen dataset. On the other hand, our proposed multi-modal learning-based method alleviates these problems.

In order to improve the performance of semantic mapping that can be obtained from a single modality (e.g., RGB or IRRG), additional modalities, either from the same sensor (e.g. multi-spectral or hyperspectral images) or from a different one (e.g., LiDAR point cloud data or SAR) are increasingly used for land cover mapping~\citep{hazirbas2016fusenet, xu2017multisource, audebert2018beyond}. Examples include Synthetic Aperture Radar (SAR) images~\citep{hong2020more, li2020multimodal}, hyperspectral imagery (HSI)~\citep{xu2017multisource, audebert2019deep} and Digital Surface Models (DSM)~\citep{hazirbas2016fusenet, audebert2018beyond}. Multi-modal data has been proven to provide rich complementary information to deal with complex scenes as different imaging technologies in RS are capable of capturing a variety of properties from the earth's surface, such as height information, spectral radiance, and reflectance~\citep{gomez2015multimodal}. 

One of the main challenges in the utilization of multi-modal data is how to effectively extract and fuse multi-modal features. Although deep learning-based methods can automatically learn representative features, multi-modal inputs and features often provide unequal, redundant, or even contradictory information. Current multi-modal models tend to extract features independently using two separate encoders, combining feature maps indiscriminately at early and/or late layers via concatenation or summation~\citep{couprie2013indoor,audebert2018beyond}. We argue that this design leads to both inaccurate and computationally inefficient models. In particular, it brings high sensitivity to missing or noisy data~\citep{audebert2018beyond}, which has a significant negative influence on overall model performance when dealing with missing or noisy modality scenarios~\citep{michael2018}. Another challenge of pixel-wise classification of multi-modal images is the increased model size and computational burden~\citep{MarmanisSWGDS16, audebert2018beyond} that also limit the application in most scenarios with real-time requirements. Hence, the effective and efficient fusion of multi-modal information is still an open research direction and also needs to be further optimized for scalability and real-time consideration for real-world applications.

Recently, the usage of attention mechanisms and graph-based approaches has led to promising performance and computational efficiency gains on a range of different computer vision tasks~\citep{mou2019learning,fu2019dual,Liu_2020_CVPR_Workshops}.
These works, typically use these mechanisms to emphasize salient features and suppress irrelevant signals in unimodal settings. Further, they tend to ignore multi-scale information by only leveraging same-dimensional representations of the same scale (e.g., typically low-spatial-resolution feature spaces) in order to alleviate the computational cost. To facilitate an efficient multi-scale (pyramid) attention feature extraction from each modality, we propose a pyramid attention fusion (PAF) module for extracting multiple hierarchical-scale representations. By using a novel gated fusion unit (GFU) to blend complementary features between multi-modal encoders, 
we introduce a lightweight multi-modal segmentation network (MultiModNet). For more details, please refer to Section~\ref{method}.

Our experiments demonstrate that the network achieves robust and accurate results on the representative ISPRS Semantic Labeling Contest Vaihingen dataset~\citep{isprs2012} and the Agriculture-Vision challenge dataset~\citep{chiu2020agriculturevision}. The main contributions of this paper are as follows: 
\begin{enumerate}
\item We present a novel pyramid attention and gated fusion mechanism for multi-modality data that builds on our proposed gated fusion unit (GFU) and our pyramid attention fusion (PAF) module. It facilitates interactions between the encoders of each modality to effectively combine the extracted features from multiple modalities and weaken the influence of noise and redundancies among the multi-modal data. 
\item The proposed PAF module is a lightweight network with a built-in cross-hierarchical-scale and cross-view attention fusion mechanism that can obtain rich and robust contextual representations. It can be used as a stand-alone decoder for a unimodal model to improve segmentation performance, or as a vital fusion mechanism to merge several modalities when combined with our gated fusion unit.
\item Built upon the PAF and GFU modules, our end-to-end multi-modal segmentation model (MultiModNet) achieves state-of-the-art performance and outperforms the baselines on two representative remote sensing datasets with considerably fewer parameters and at a lower computational cost. We also validate the effectiveness and flexibility of our framework through extensive ablation studies. 
\end{enumerate}

The paper is structured as follows. Section~\ref{related} provides an overview of the related work. In Section~\ref{method}, we present the methodology in detail. Experimental procedure and evaluation of the proposed method is performed in Section~\ref{exp}. Section~\ref{addtion} further discusses and evaluates our method via ablation studies. Finally, we draw conclusions and outline future research directions in Section~\ref{concl}.

\section{Related work}
\label{related}
The state-of-the-art deep learning-based segmentation models are mostly inspired by the idea of fully convolutional networks (FCNs)~\citep{long2015fully}. FCN models generally consist of an encoder-decoder architecture in which all layers are based on convolutions (and upsampling/downsampling operations). However, vanilla FCNs tend to cause a loss of spatial information due to the presence of pooling layers that reduce the resolution of feature maps by sacrificing the positional information of objects. The UNet~\citep{ronneberger2015u} extends the FCN by introducing symmetric skip connections (i.e., concatenations) between the encoder and decoder modules to maintain spatial information. The precise spatial information can be gradually recovered in the decoder module by combining multiple skipped connections with upsampling or de-convolution layers. Since then, the encoder-decoder architecture has been widely extended in recent works including, among others, pyramid scene parsing network (PSPNet)~\citep{zhao2016pyramid}, SegNet~\citep{badrinarayanan2017segnet}, DeepLabV3+~\citep{chen2018deeplab}, Dual attention network~\citep{fu2019dual}, and HRNet-OCR~\citep{YuanCW19}. 

The FCN-based or UNet-based encoder-decoder architectures have also been widely adopted and applied to the ISPRS Semantic Labeling Contest~\citep{paisitkriangkrai2015effective, kampffmeyer2016semantic, Sherrah16, lin2016efficient, MarmanisSWGDS16, audebert2016semantic, audebert2017joint, wang2017gated, michael2018, liu2020TGRS}, and the Agriculture-Vision benchmark dataset for automatic mapping of land pattern types ~\citep{chiu2020agriculturevision, Liu_2020_CVPR_Workshops, liu_00}. In general, these architectures differ from each other in how they capture rich and global contextual information at multiple scales. For instance, the stacked UNet architecture is proposed by \cite{ghosh2018stacked} for land cover segmentation in remote sensing imagery, which merges high-resolution details and long range contextual information captured at low-resolution to generate segmentation maps. Further,~\cite{liu2020TGRS} introduced a dense dilated convolutions merging (DDCM) network that sequentially stacked the output of each layer with its input features before feeding it to the next layer to capture global and multi-scale contextual features. 

Despite the aforementioned impressive progress on unimodal deep learning, deep learning has also been exploited for multi-modal data processing to obtain finer representations of different modalities. From the perspective of multi-modal fusion in remote sensing, a multi-modal deep learning model normally involves concatenation of extracted features from unimodal networks (e.g., a backbone network) and then learning a joint representation for classification or segmentation. A representative work proposed by~\cite{audebert2018beyond} investigated two fusion strategies, namely early and late fusion methods based on the FuseNet framework, using SegNet or ResNet to classify multi-modal remote sensing data (such as LiDAR and multispectral images). Specifically, one CNN-based encoder (e.g, VGG or ResNet) is used to extract the features from RGB or IRRG images while another encoder is exploited to extract the features from LiDAR data and other bands (e.g NDVI). Note that the LiDAR data has been rasterized in the image domain as a digital surface model (DSM) with normalization (nDSM). Early fusion concatenates the features after each convolutional block from both encoders, while later fusion merges the last feature maps from the two deep networks. The results show that late fusion improves the overall accuracy at the cost of less balanced predictions, while early fusion achieves better performance for all classes but inducing higher sensitivity to missing or noisy data. Indeed, such fusion techniques do require all modalities to be available to the classification during both training and testing. \cite{michael2018} therefore presented a novel CNN architecture based on so-called hallucination networks for urban land cover classification that can replace missing data modalities in the test phase. This enables fusion capabilities even when data modalities are missing in testing. Lately, \cite{feng2019multisource} presented an adaptive approach to fuse HSI and LiDAR data, in which a two-stream CNN is used to extract LiDAR and HSI features separately. Then an adaptive method based on squeeze-and-excitation networks~\citep{hu2018squeeze} is designed to combine the features with adaptive weights instead of simply concatenation. \cite{fusion_xu} and \cite{fusion_contest} further proposed a Fusion-FCN framework for the classification of multi-source remote sensing data using fused FCNs where three different types of data (LiDAR data, hyperspectral images, and very high-resolution RGB images) are utilized in one model.

Recently, the usage of attention mechanisms in deep learning models has been increasingly explored in various visual inference tasks and has shown very promising performance gain~\citep{DANet,mou2019learning,FusAtNet}. Generally, the attention modules highlight the prominent features while suppressing the irrelevant features through a self-attention learning method~\citep{Attentionisall}. Recent work by \cite{Liu_2020_CVPR_Workshops} proposed a multi-view graph-based attention paradigm (MSCG) that demonstrated significant performance gain in contrast to a single-view attention module (SCG)~\citep{liu2020scg} for land cover mapping of multi-spectral aerial images. However, in most of these works, the attention modules are carried out only on single-level features with coarse resolution from a single modality to alleviate the computational cost. This brings challenges when attempting to accurately classify relatively small objects in very high-resolution remote sensing data. To alleviate these problems, our focus in this paper is mainly on land cover mapping (pixel-wise classification) tasks of multi-modal remote sensing images, facilitated through our proposed multi-scale and cross-view attention fusing mechanism.

\section{Method}
\label{method}
The proposed multi-modality network (MultiModNet) consists of four key modules: a backbone encoder (ENC), a pyramid attention fusion (PAF) module, a gated fusion unit (GFU), and a decoder (DEC) that produces the final output. 
\begin{figure}[htbp]
 \centering
  \includegraphics[width=0.85\columnwidth]{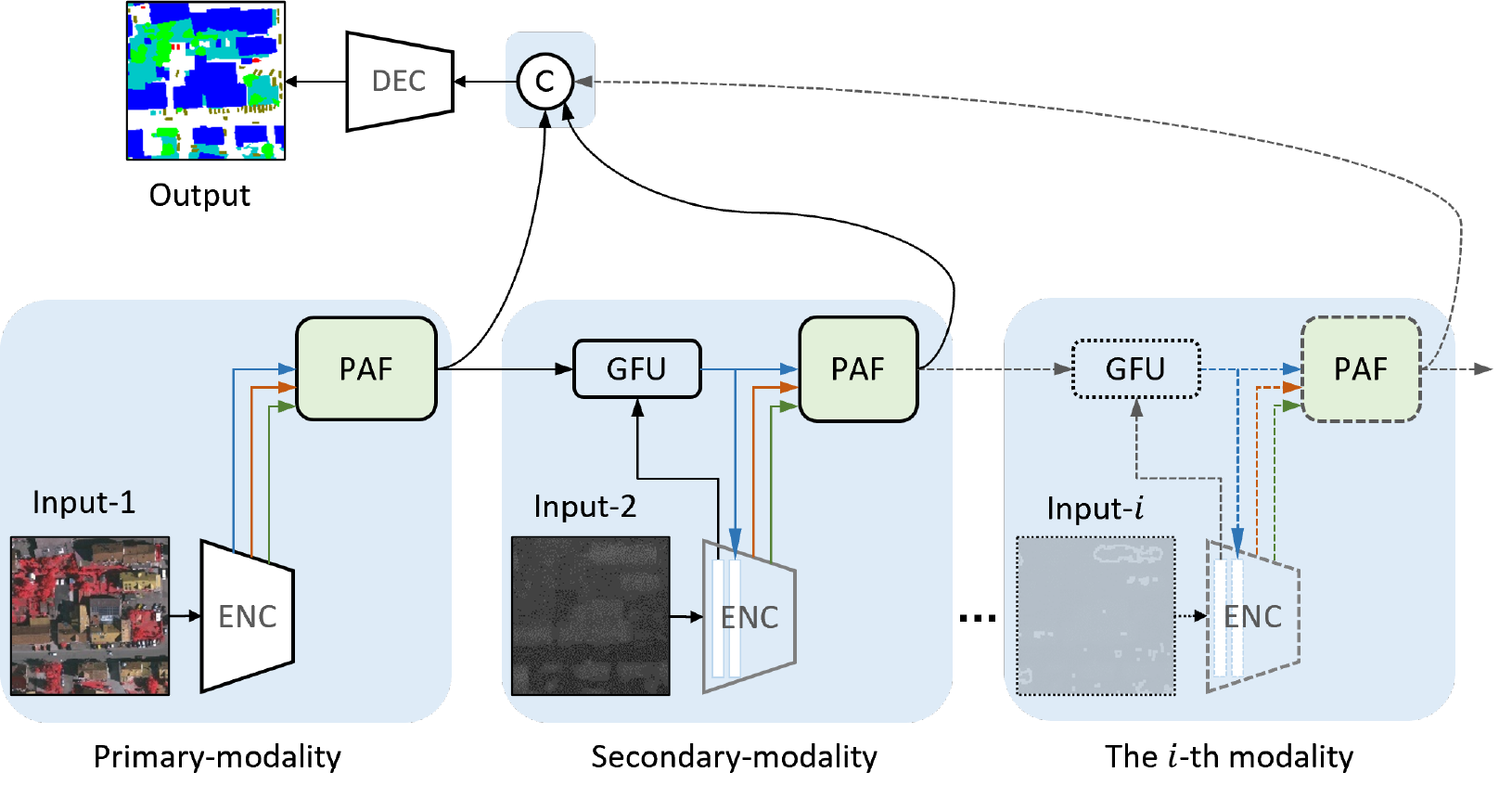} 
  \caption{General concept structure of our multi-modality network (MultiModNet) based on proposed pyramid attention and gated fusion methods. Here ENC denotes the feature encoder, {\text{GFU}} accounts for a gated fusion unit, {\text{PAF}} is our proposed pyramid attention fusion module, {\textcircled{c}} denotes concatenation, Input-1, Input-2 and $\text{Input-}i$ are the primary, the secondary and the $i$-th modalities respectively, and, DEC is the decoder layer to output the final prediction. Note that our PAF module normally takes three different scale (i.e. 3-level) features of each modality as the input shown with blue, orange and green line.}
  \label{fig:hpaf_arch}
\end{figure}
Given a primary and a secondary modality\footnote{There will be a third or even more supplementation modalities, we thus describe them using the $i$-th modality as illustrated in Figure 2, and assume they are ordered depending on informational richness and significance, i.e., Input-$1$ $\geq$ Input-$2$ $\geq$ $\cdots$ $\geq$ Input-$i$. In other words, each preceding modality can be seen as a primary modality with respect to the following (succeeding, if any) ones. } Input-1 and Input-2, respectively, where Input-1, e.g. IRRG or RGB images, contains more valuable information than Input-2 , e.g. DSM or NIR images, off-the-shelf encoders (ENC), such as multi-layer CNN based backbones (e.g. ResNet), are used to extract multi-level feature maps for each modality. Then we utilize PAF modules (Section \ref{paf}) to generate fine-grained cross-level features and GFUs (Section \ref{gfu}) to merge complementary features from the primary modality into the secondary modality. Finally, we concatenate all PAF generated features from each modality and feed them into a simple decoder (DEC) module, which in this work is composed of only a single convolution layer and a bi-linear interpolation function, to output the pixel-wise classification maps. As shown in Fig.~\ref{fig:hpaf_arch}, our MultiModNet framework has a scalable structure that allows it to easily extend to more than two modalities. The parts that follow will go through our PAF and GFU modules in detail.


\subsection{Pyramid Attention Fusion}\label{paf}
\begin{figure}[htbp]
 \centering
  \includegraphics[width=0.9\columnwidth]{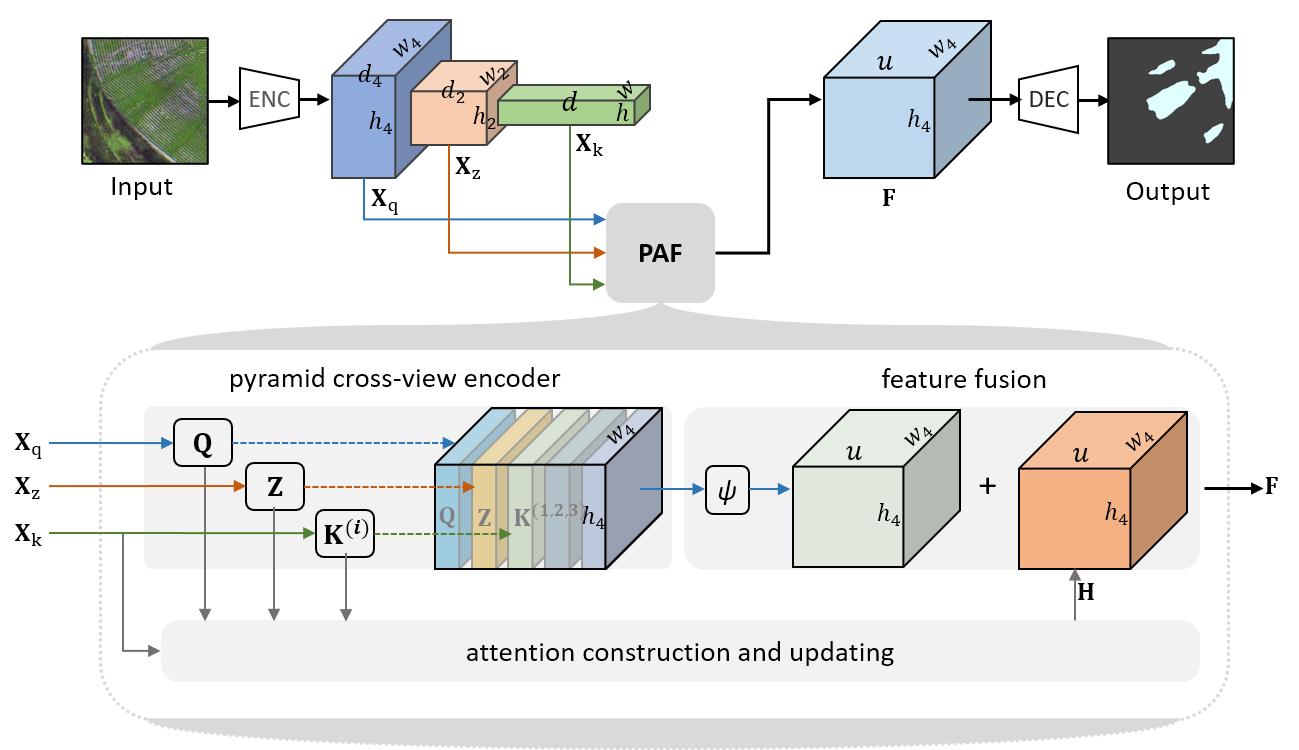}
  \caption{The illustration diagram of the pyramid attention fusion (PAF) module. Overall PAF is composed of three key blocks, i.e., the \textit{pyramid coss-view encoder} that transforms the input pyramid features (i.e., $\textbf{X}_{\text{q}}$, $\textbf{X}_{\text{z}}$ and $\textbf{X}_{\text{k}}$, obtained by the ENC module) to corresponding multi-scale latent spaces (i.e., $\textbf{Q}$, $\textbf{Z}$ and $\textbf{K}^{(i)}$), the \textit{attention construction and updating} block that constructs the cross-view attention matrix and then transforms the high-level features onto high-resolution 2D attention features ($\textbf{H}$) by a message-passing function (see Eq. \ref{eq:adj_update}), and the \textit{feature fusion} unit, which combines the latent multi-scale features with a CNN network ($\psi$) and sums the learned attention features ($\textbf{H}$) to eventually generate the fused feature map ($\textbf{F}$), which can then be fed into the DEC module to produce the final output.}
  \label{fig:ae_encoder}
\end{figure}
We develop the lightweight PAF module with a built-in cross-hierarchical-scale and cross-view attention fusion mechanism that can obtain rich and robust  representations. The features produced from the PAF module at each previous modality will be integrated into the encoder layer of its successor modality through a GFU module. The proposed PAF module thus plays a vital role in fusing a range of modalities in a compact yet effective manner, while it can still be used as a stand-alone decoding layer for unimodal models to improve segmentation performance. 

As illustrated in Fig.~\ref{fig:ae_encoder}, our PAF module contains three key sub-blocks: the pyramid cross-view encoder, the attention construction and updating block, and the feature fusion block:
\begin{itemize}
    \item The \textit{pyramid cross-view encoder} transforms the selected three-different-size feature maps (e.g., $\textbf{X}_{\text{q}}$, $\textbf{X}_{\text{z}}$, $\textbf{X}_{\text{k}}$) to corresponding cross-level and cross-view latent representations (i.e., $\textbf{Q}$, $\textbf{Z}$ and $\textbf{K}^{(i)}$), in order to decrease computational burden while extracting salient latent features for late cross-view attention map generation. 
    \item The \textit{attention construction and updating} block constructs the cross-view and cross-level attention matrix (see Eq. \ref{eq:adj_construct}) and then transforms the high-level features onto high-resolution 2D attention representations $\textbf{H}$ by a message-passing function (see Eq. \ref{eq:adj_update}), in order to obtain robust non-local and high-resolution contextual features. 
    \item The \textit{feature fusion} module combines the latent cross-level and cross-view features with a CNN network (denoted by $\psi$) and sums the learned high-resolution attention representation $\textbf{H}$, in order to eventually produce the fine-grained contextual features $\textbf{F}$, which can then be fed into the DEC module to produce the final output.
\end{itemize}

We describe each component of the framework in detail as follows.

\subsubsection{Pyramid cross-view encoder}
To reduce computational cost while obtaining robust latent feature representations for late constructing attention maps, we utilize a multi-view augmenting method \citep{Liu_2020_CVPR_Workshops} in the pyramid cross-view encoder to explicitly exploit the rotation invariance in the deep features. We first define a view generation function $\textbf{X}_{\text{k}}^{(i)} = \tau(\textbf{X}_{\text{k}}, i)$, and a view reversion function $\textbf{X}_{\text{k}} = \tau^{-1}(\textbf{X}_{\text{k}}^{(i)}, i)$ for three different views ($i \in \{1, 2, 3\}$). We let $\textbf{X}_{\text{k}}^{(1)}=\textbf{X}_{\text{k}}$, and generate $\textbf{X}_{\text{k}}^{(2)}$ and $\textbf{X}_{\text{k}}^{(3)}$ by transposing and vertically flipping, respectively. Then the module learns pyramid-level and cross-view latent representations, i.e, a low-level feature matrix $\textbf{Q} \in \mathbb{R}^{h_4 \times w_4 \times c}$, a middle-level latent matrix $\textbf{Z} \in \mathbb{R}^{h_2 \times w_2 \times c}$ and the high-level 3-view matrix $\textbf{K}^{(i)} \in \mathbb{R}^{h \times w \times c}$ from the multi-scale features $\textbf{X}_{\text{q}} \in \mathbb{R}^{h_4 \times w_4 \times d_4}$, $\textbf{X}_{\text{z}}\in \mathbb{R}^{h_2 \times w_2 \times d_2}$, and $\textbf{X}_{\text{k}}\in \mathbb{R}^{h \times w \times d}$, respectively, using CNNs, i.e., 
\begin{equation}\label{eq:qzk}
   \textbf{Q}=\varphi\left(\textbf{X}_{\text{q}} ; \boldsymbol{\theta}_{\text{q}}\right), \quad \textbf{Z}=\varphi\left(\textbf{X}_\text{z} ; \boldsymbol{\theta}_{\text{z}}\right), \quad \text{and} \quad \textbf{K}^{(i)}=\tau^{-1}\left(\varphi\left(\textbf{X}_{\text{k}}^{(i)} ; \boldsymbol{\theta}_{\text{k}}\right), i\right)\;,
\end{equation}
where $\varphi$ denotes the convolution layers with parameter kernels of $\boldsymbol{\theta}_{\text{q}} \in \mathbb{R}^{d_4 \times 3 \times 3 \times c}$, $\boldsymbol{\theta}_{\text{z}}\in \mathbb{R}^{d_2 \times 3 \times 3 \times c}$, and $\boldsymbol{\theta}_{\text{k}} \in \mathbb{R}^{d \times 3 \times 3 \times c}$ respectively. Note that $d_4$, $d_2$, and $d$ represent the input feature dimensions of $\textbf{X}_{\text{q}}$, $\textbf{X}_{\text{z}}$ and $\textbf{X}_{\text{k}}$  respectively, $c$ is the output feature dimension, and typically $c < d_4 < d_2 < d$. Here, $h_4\times w_4$, $h_2\times w_2$ and $h\times w$ denote the spatial sizes of both the input and the output feature maps, and commonly $h_4 = 2h_2 = 4h$, $
w_4 = 2w_2 = 4w$. We also use zero-padding methods in CNN layers of the module to keep the output spatial resolution the same as the input.

\subsubsection{Attention construction and updating}
To obtain a robust non-local and high-resolution contextual feature space based on these learned three-level and three-view latent representations (i.e., $\textbf{Q}$, $\textbf{Z}$, $\textbf{K}^{(i)}:i=1,2,3.$), we propose a novel attention construction and updating module that can efficiently model long-range and cross-level pixel-wise dependencies and effectively produce rich non-local and high-resolution contextual representations via an upsampling-based attention-passing mechanism. This module is formed of two key components: attention construction and attention-passing. They are described in detail as follows.

\begin{figure}[htbp]
 \centering
  \includegraphics[width=0.90\columnwidth]{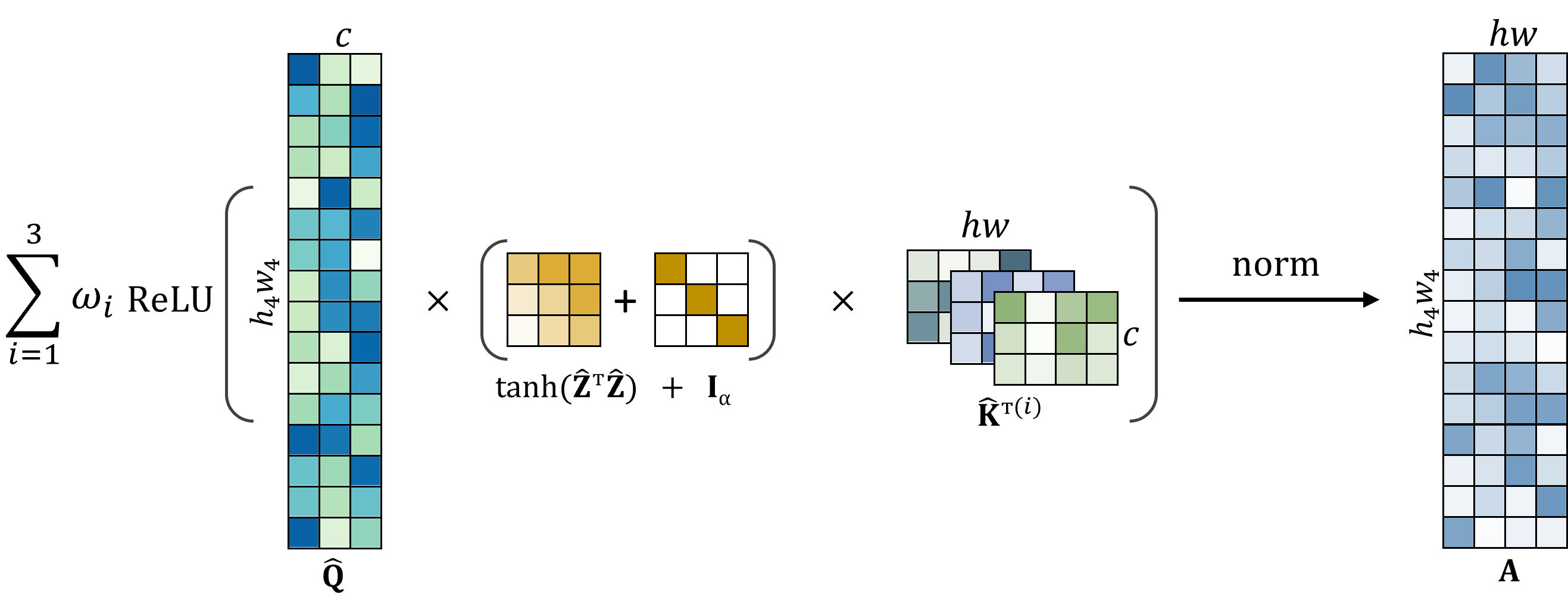}
  \caption{The illustration of attention construction}
  \label{fig:att_constr}
\end{figure}

\textbf{Attention construction}. Inspired by the success of self-attention \citep{Attentionisall} to encode the structural information of a sequence of data, we present a long-range cross-level attention method that uses latent feature similarity to model the interactions between every pair of pixels in cross-level feature maps. Furthermore, we introduce a multi-view fusion strategy in the attention module, which allows us to encode cross-level as well as cross-view pixel-wise dependencies to improve its robustness. Specifically, we first reshape the low-level high-resolution latent matrix $\textbf{Q}$ to $\hat{\textbf{Q}} \in \mathbb{R}^{(h_4w_4)\times c}$, the middle-level latent matrix $\textbf{Z}$ to $\hat{\textbf{Z}} \in \mathbb{R}^{(h_2w_2)\times c}$ and the high-level view matrices $\textbf{K}^{(i)}$ to $\hat{\textbf{K}}^{(i)} \in \mathbb{R}^{(hw)\times c}$. Then our cross-view and cross-level attention construction function is defined as
\begin{equation}\label{eq:adj_construct}
\textbf{A} = \operatorname{norm} \left(\overbrace{\sum_{i=1}^{3} w_{i}}^{\substack{\text{cross-view} \\ \text{fusion}}} \operatorname{ReLU}\left(\overbrace{\hat{\textbf{Q}} \left(\underbrace{\operatorname{tanh}\left(\hat{\textbf{Z}}^{\top}\hat{\textbf{Z}}\right) + \textbf{I}_\alpha}_{\text{channel-wise attention}}\right)\hat{\textbf{K}}^{\text{T}(i)} }^{\text{long-range cross-level attention}}\right)\right) \in \mathbb{R}^{(h_4w_4) \times (hw)}
\end{equation}
where $w_i$ is a learnable parameter initialized as 1 for our attention construction function, $\operatorname{tanh}(\cdot)$ and $\operatorname{ReLU}(\cdot)$ denote the $\operatorname{tanh}$ and $\operatorname{ReLU}$ non-linear functions respectively, and $\textbf{I}_\alpha \in \mathbb{R}^{c \times c}$ is a learnable bias kernel initialized as $\mathbf{I}$. Note that the attention matrix $\textbf{A}$ is constructed from features from different scales, resulting in long-range cross-level attention. The matrix is therefore a tall matrix, i.e. it has more rows (e.g., $16hw$) than columns (e.g., $hw$), and it is further normalized, i.e. $\operatorname{norm}(\cdot)$, along rows by dividing by the sum of each row, so that the elements of each row vector in the matrix add up to 1. Figure \ref{fig:att_constr} illustrates the attention matrix constructing process.

Note that our attention construction process differs from the self-attention scheme in three major ways, i.e.,
\begin{itemize}
    \item \textit{Cross-level attention:} Our cross-level attention scheme utilizes three distinct level feature maps as the sources of multi-scale latent representations to efficiently generate a tall non-local interaction matrix instead of a square self-attention matrix. This allows our attention module to learn high-resolution features from low-resolution but high-abstract feature space. Based on our observations, using our cross-level attention to capture contextual information leads to faster training and better performance on remote sensing data than using self-attention methods based on one-scale low-resolution features or image patches \citep{dosovitskiy2020vit}.
    \item \textit{Channel-wise attention:} We also integrate a channel-wise attention method, i.e., $\operatorname{tanh}\left(\hat{\textbf{Z}}^{\top}\hat{\textbf{Z}}\right) + \textbf{I}_\alpha$, into our long-range cross-level attention scheme (see Eq. \ref{eq:adj_construct}) to improve feature discriminability by blending channel-wise weights learned from the middle-level feature ($\textbf{Z}$) space. We observe that this results in better training stability and less sensitivity to latent feature dimensionalities (i.e., $c$) when compared to not using the channel-wise attention mechanism. We think that the channel-wise attention, like the dual attention network \citep{DANet}, could enhance our long-range attention mechanism by merging both channel and spatial attention attributes to capture robust cross-level information.
    \item \textit{Cross-view fusion:} Furthermore, we introduce a cross-view fusion strategy into our attention module, inspired by our previous work \citep{Liu_2020_CVPR_Workshops}, to explicitly encode the rotation invariance in the high-abstract and deep-level latent features (i.e., $\textbf{K}$).
    We fuse (add up) three-view long-range attention maps using learnable weights (i.e., $w_i$ in Eq. 2) to further improve the model's robustness. Base on our experiments, using cross-view attentions can further speed up the model's learning process and result in better performance than using single-view attention maps.
\end{itemize}

\begin{figure}[htbp]
 \centering
  \includegraphics[width=0.95\columnwidth]{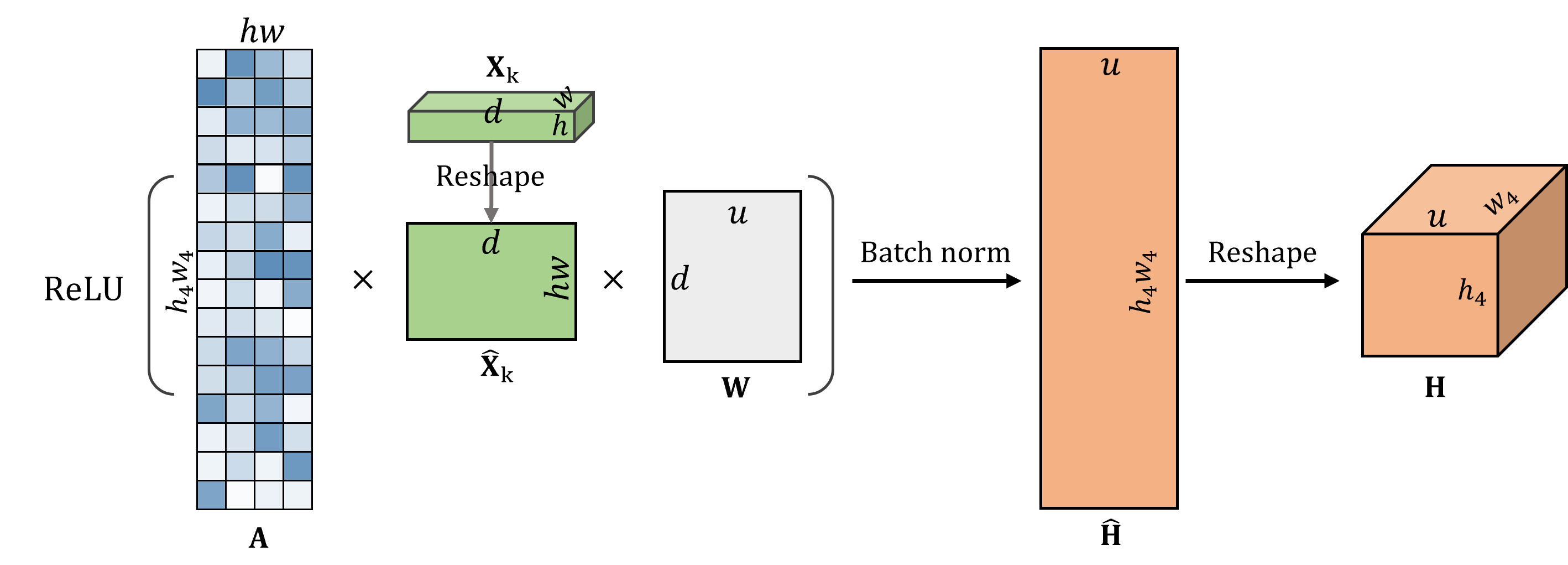}
  \caption{The illustration of attention-passing pipeline.}
  \label{fig:att_update}
\end{figure}

\textbf{Attention-passing.} To produce a non-local but high-resolution feature representation (i.e., $\textbf{H}$: typically 4 times the size of $\mathbf{X}_{\text{k}}$) from the high-level but low-resolution features $\mathbf{X}_{\text{k}}$, we develop an upsampling-based attention-passing function $f(\cdot )$ (Eq. \ref{eq:adj_update}). It is parameterized by the normalized attention matrix $\textbf{A}$ and $\mathbf{X}_{\text{k}}$ with trainable parameters $\textbf{W} \in \mathbb{R}^{d \times u}$ where $u$ denotes the output feature dimension. Our attention-passing mechanism, i.e., $\textbf{A}\hat{\textbf{X}}_{\text{k}}\textbf{W}$, is similar to the one-hop neighborhood message-passing function of graph convolutional networks~\citep{kipf2016semi} when viewing our learned tall attention matrix as a special type of adjacency matrix. Note that $\hat{\textbf{X}}_{\text{k}} \in \mathbb{R}^{(hw) \times d}$ is obtained by reshaping of $\textbf{X}_{\text{k}}$. 
\begin{equation}\label{eq:adj_update}
    \begin{aligned}
\mathbf{H} &= f\left(\mathbf{X}_{\text{k}} \in \mathbb{R}^{h \times w \times d}; \mathbf{A}, \mathbf{W}\right) = \delta \left(\mathbf{A} \hat{\mathbf{X}}_{\text{k}}\textbf{W}\right) \in \mathbb{R}^{h_4 \times w_4 \times u} \;.
\end{aligned}
\end{equation}
With a combination operator, denoted by $\delta(\cdot)$ in Eq. \ref{eq:adj_update}, of non-linear activation function (e,g. $\operatorname{ReLU}$) with bath normalization  and reshaping, we eventually obtain a high-resolution attention representation $\textbf{H} \in \mathbb{R}^{h_4 \times w_4 \times u}$ as illustrated in Figure \ref{fig:att_update}.

\subsubsection{Feature fusion} 
Finally, we fuse the high-resolution attention features with cross-level and cross-view latent features in order to produce fine-grained high-resolution representations with robust non-local contextual and spatial information as the output using
\begin{equation}\label{eq:aux_p}
\mathbf{F} = \psi \left( \textbf{Q} \parallel \tilde{\textbf{Z}} \parallel \tilde{\textbf{K}}^{(1)} \parallel \tilde{\textbf{K}}^{(2)} \parallel \tilde{\textbf{K}}^{(3)}; \boldsymbol{\theta}_{\psi}\right) + \textbf{H}\;,
\end{equation}
where $\psi$ denotes the convolution layer with parameter kernels of $\boldsymbol{\theta}_{\psi} \in \mathbb{R}^{5c \times 3 \times 3 \times u}$, batch normalization and non-linearity, and $\parallel$ denotes concatenation.
Please note that middle-level latent and high-level view feature matrices ($\textbf{Z}\text{ and }\textbf{K}^{(i)}$) are up-sampled using bi-linear interpolation to $\tilde{\textbf{Z}}\text{ and }\tilde{\textbf{K}}^{(i)}$ in order to match the dimension of the high-resolution (i.e. $h_4\times w_4$) feature matrix $\textbf{Q}$ for concatenating. This is similar to the multi-level feature fusion method of pyramid feature networks (FPNs) \citep{lin2017feature}, which fuse multi-level features from the top-down path by upsampling and summing, but instead of summation, we use concatenation and convolution operations to merge multi-level feature maps for remote sensing data.

\subsection{Gated Fusion Unit} \label{gfu}
\begin{figure}[htbp]
 \centering
  \includegraphics[width=0.6\columnwidth]{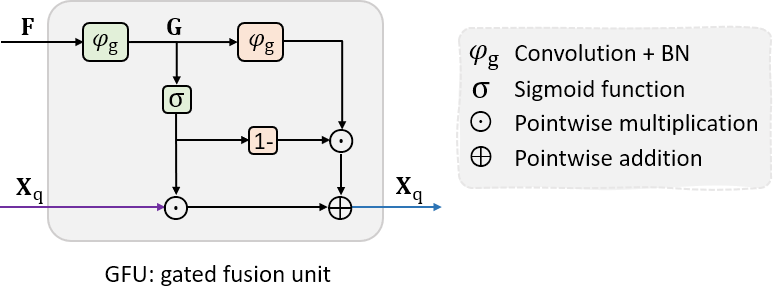} 
  \caption{The gated fusion unit (GFU) consists of 2 CNN layers ($\varphi_\text{g}$) with batch normalization, and an activation function ($\sigma$, i.e,  Sigmoid), where '1-' denotes one minus the input activation maps.}
  \label{fig:gfu_arch}
\end{figure}

The GFU module is designed to serve as a fusion gateway between the main and secondary modalities. It utilizes a novel gating mechanism to allow the primary modality to aid its secondary modality in extracting the supplementary information via a gating network, thereby minimizing the influence of hidden noise and redundancies.
Specifically, the GFU module is composed of two CNN layers with two gating operations (element-wise multiplications) as shown in Fig.~\ref{fig:gfu_arch}. The first gate operation helps to weaken redundancies and capture salient useful features from the secondary modality, while the second gate operation aims to obtain complementary features from the primary modality and merge them into the secondary modality. 
The operation of the GFU module can be summarized by the following mathematical equations:
\begin{equation}\label{eq:gfu}
   \textbf{G}=\varphi_\text{g}\left(\textbf{F}; \boldsymbol{\theta}_{\text{s}}\right), \qquad \textbf{X}_\text{q}= \sigma\left(\textbf{G}\right) \odot \textbf{X}_\text{q} + \left(1 -\sigma\left(\textbf{G}\right) \right) \odot \varphi_\text{g}\left(\textbf{G}; \boldsymbol{\theta}_{\text{r}}\right) \;,
\end{equation}
where $\textbf{F}$ represents the fused representations by the PAF module of the primary modality and $\textbf{X}_\text{q}$ denotes the low-level features extracted by the encoder of the secondary modality. Here $\varphi_\text{g}$ represents the convolution layers with $1 \times 1$ filters $\boldsymbol{\theta}_\text{s}$ and $\boldsymbol{\theta}_\text{r}$ respectively, and combine a batch normalization operator. $\sigma$ is a sigmoid activation function. Note that the updated $\textbf{X}_\text{q}$ (the output of GFU) will feed the remaining layers of the encoder and also serve as one of the three input feature maps to the PAF module.

\section{Data, experiments and results}
\label{exp}
\subsection{Benchmark datasets}
In this paper, we focus on two different representative databases, namely the ISPRS Vaihingen 2D dataset~\citep{isprs2012} and the Agriculture-Vision\footnote{https://www.agriculture-vision.com/agriculture-vision-2021/dataset-2021} challenge dataset~\citep{chiu2020agriculturevision}. The ISPRS Vaihingen 2D dataset\footnote{http://www2.isprs.org/commissions/comm3/wg4/2d-sem-label-vaihingen.html} is comprised of aerial remote sensing images over the city Vaihingen in Germany. The Agriculture-Vision dataset consists of large-scale high-quality aerial images from $3,432$ farmlands across the US and has been annotated with nine types of field anomaly patterns that are most important to farmers. Each dataset provides online leaderboards and reports test metrics measured on hold-out test images. 
\begin{figure}[htpb!]
 \centering
  \includegraphics[width=0.8\textwidth]{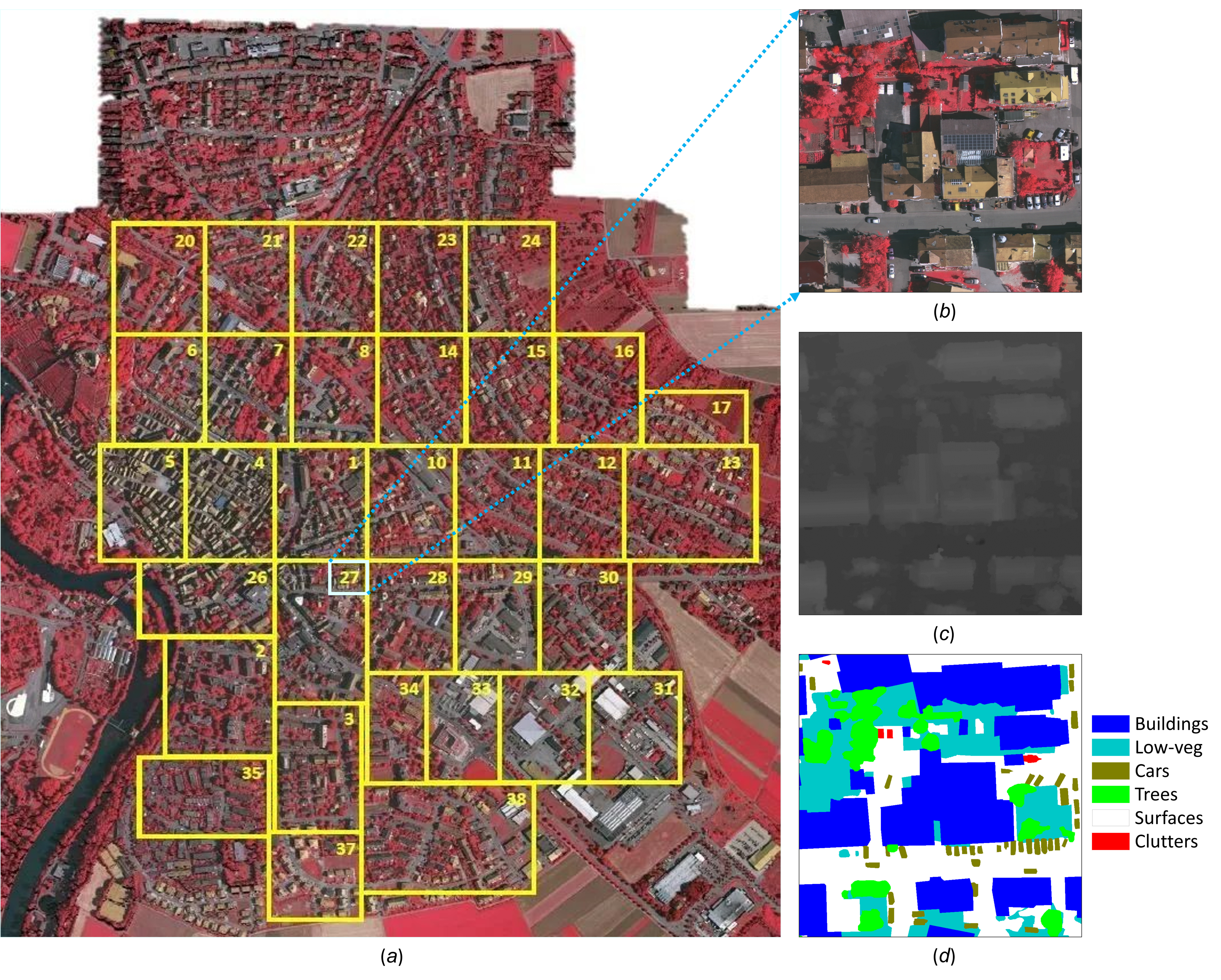}
  \caption{Overview of the ISPRS Vaihingen 2D semantic labeling benchmark dataset that contains 33 tiles: (a) overview of the entire dataset (the ID number labeled in the upper right corner of each area), (b) the IRRG image patch, (c) the DSM, (d) the ground truth.}
  \label{fig:datasets}
\end{figure}

\subsubsection{Vaihingen dataset}
The Vaihingen dataset is composed of 33 orthorectified image tiles acquired by a near-infrared (NIR) - red (R) - green (G) aerial camera and has been labeled with six common land cover categories: impervious surfaces (i.e., roads and concrete surfaces), buildings, low vegetation, trees, cars and clutter (representing uncategorizable land covers). 16 out of the 33 tiles are fully annotated at pixel level as the training set, and 17 tiles (i.e., areas: 2, 4, 6, 8, 10, 12, 14, 16, 20, 22, 24, 27, 29, 31, 33, 35 and 38) are used as hold-out test images as shown in Fig.~\ref{fig:datasets}. The average size of the tiles is approximately $2500 \times 2000$ pixels with a ground resolution of 9cm. 

Images are accompanied by a digital surface model (DSM) that is derived from dense image matching techniques and represents the absolute height of pixels. Normalized DSM (nDSM) data are also included, which represent the pixels heights relative to the elevation of the nearest ground surface. We use both IRRG and nDSM data for training and test. Fig.~\ref{fig:datasets} shows some examples of the dataset. 
\begin{figure}[htbp]
 \centering
  \includegraphics[width=1\columnwidth]{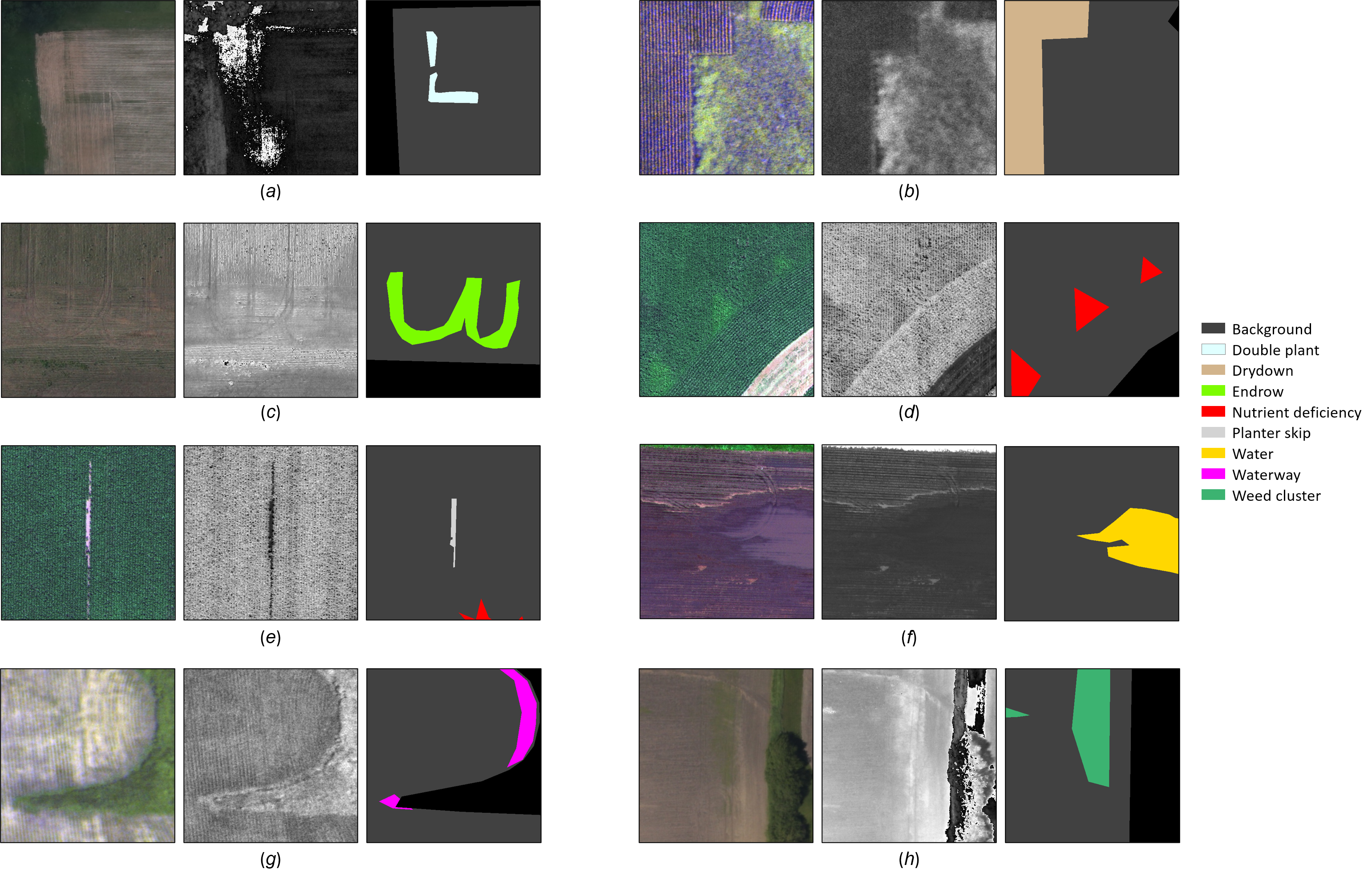} 
  \caption{Some image examples of the agriculture-vision dataset: (a) the double plant patches (left: RGB image, middle: NIR image, right: the ground truth), (b) the drybown patches, (c) the endrow patches, (d) the nutrient deficiency patches, (e) the planter skip patches, (f) the water patches, (g) the waterway patches, (h) the weed cluster patches.}
  \label{fig:data_example}
\end{figure}

\subsubsection{Agriculture-Vision dataset}
The Agriculture-Vision dataset consists of $94,986$ aerial farmland images, of which $19,708$ images are used as the hold out test set, and $18,334$ are used as the local validation set. Each image consists of $512 \times 512$ RGB and NIR channels with resolution as high as $10$ cm per pixel. Nine types of the most important field  patterns are annotated: double plant, drydown, endrow, nutrient deficiency, planter skip, storm damage, water, waterway, and weed cluster. In addition, each image has a boundary map that indicates the region of the farmland, and a mask that indicates valid pixels in the image. Fig.~\ref{fig:data_example} shows some examples of the dataset (note that the black regions in the ground truth denote invalid areas). 

Due to the fact that some annotations may overlap in the dataset, for pixels with multiple labels, a prediction of either label will be counted as a correct pixel classification for that label. Therefore, the conventional mean Intersection-over-Union (mIoU) metric is modified accordingly by categorizing predictions of any label in a pixel as a correct prediction. This customized mIoU is used as the main quantitative evaluation metric of the contest dataset~\citep{liu_00}. 

\subsection{Variants of MultiModNet} \label{archit}
Built upon PAF, GFU, and the incorporation of different backbone encoders, we develop various MultiModNet models for land cover mapping tasks on different remote sensing data as shown in Table \ref{tab:parameters}.
\begin{table}[hptb!]
\tbl{Detailed configurations of variants of MultiModNet with quantitative comparison of parameters size, FLOPs (measured on input image size of $4 \times 512 \times 512$), Inference time on CPU and GPU separately.}
 {\resizebox{\columnwidth}{!}{
\begin{tabular}{lccccccccc} \hline 
\thead{Models} & \thead{Inputs} & \thead{ENC-1} & \thead{ENC-2} & \thead{PAF-1} & \thead{PAF-2} & \thead{GFU} & \thead{Parameters \\ (Million)}& \thead{FLOPs \\ (Giga)}& \thead{Inference time \\ (ms - CPU/GPU)}   \\ \midrule   
$\text{PAFNet}^{a}$ &\thead{DSM-IRRG \\ (unimodal 4-band)} &\thead{Se\_ResNext50 \\ (output: 256/512/1024)} & \xmark & \thead{latent:6 \\ output:24}& \xmark  & \xmark& 9.52  &18.12 & {513} / 24 \\
$\text{PAGNet}^{a}$ &\thead{IRRG, DSM \\ (Two-modal)}&\thead{Se\_ResNext50 \\ (output: 256/512/1024)} & \thead{MobileNetV3 \\ (output: 40/112/960)} &  \thead{latent:6 \\ output:24} & \thead{latent:6 \\ output:24}& \thead{output:40} & 12.62  &21.34 & {564} / 36 \\\hdashline
$\text{PAFNet}^{b}$ & \thead{NIR-RGB \\ (unimodal 4-band)} & \thead{MobileNetV3 \\ (output: 40/112/960)} & \xmark & \thead{latent:8 \\ output:32}& \xmark& \xmark & 4.86  &5.23 & {68} / 6 \\
$\text{PAGNet}^{b}$ &\thead{RGB, NIR \\ (Two-modal)} & \thead{MobileNetV3 \\ (output: 40/112/960)} & \thead{MobileNetV3 \\ (output: 40/112/960)} & \thead{latent:8 \\ output:32} & \thead{latent:8 \\ output:32} & \thead{output:40} & 6.14 &7.86 & 127 / 11  \\ 
\hline

\end{tabular}}
} 
\label{tab:parameters}%
\end{table}
Specifically, we use different backbone encoders (all are pretrained on ImageNet in this work) for Vaihingen and Agriculture-vision datasets. Our models, i.e., $\text{PAFNet}^{a}$ and multi-modal $\text{PAGNet}^{a}$ for the Vaihingen dataset, use the Se\_ResNext50~\citep{hu2018squeeze} as the ENC-1 for IRRG images and the MobileNetV3~\citep{mobilev3} as ENC-2 for DSM data respectively, while for the Agriculture-Vison dataset, the models i.e. $\text{PAFNet}^{b}$ and $\text{PAGNet}^{b}$, we use two identical MobileNetV3 models  as the encoders for both RGB and NIR data. We were not able to use Se\_ResNext50 for the Agriculture-Vision dataset due to the memory limitation (11Gb) of our GPU, since Se\_ResNext50 requires much more memory compared to MobileNetV3 when taking larger input size and batch size required for training models on the Agriculture-vision dataset. 

\subsection{Training details}
According to best practices, we train all our models using Adam~\citep{KingmaB14adam} as the optimizer for the first 10k iterations and then change the optimizer to SGD in the remaining iterations with weight decay $2 \times 10^{-5}$ applied to all learnable parameters except biases and batch-norm parameters. We use a polynomial learning rate ($\operatorname{lr}$) decay $(1 - \frac{cur\_iter}{max \_iter})^{0.9}$ with the maximum iterations set to $10^8$. We also set $2 \times \operatorname{lr}$ to all bias parameters in contrast to weights parameters. Based on our training observations to achieve fast and stable convergence, we apply the adaptive multi-class weighting loss ($\mathcal{L}_{acw}$) function~\citep{Liu_2020_CVPR_Workshops} for all our experiments.

Guided by our empirical results and our previous work~\citep{liu2020TGRS,Liu_2020_CVPR_Workshops}, we train and validate the networks for the Vaihingen dataset with 5000 randomly sampled patches of size $448\times 448$ as input and a batch size of five. For the experiments on the Agriculture-Vison dataset, we randomly sample images of size $512\times 512$ as input and train it using mini-batches of size $12$. We conduct all experiments using PyTorch on a computer with a single GeForce GTX 2080Ti. For the Vaihingen dataset, we set the initial learning rate to $1.8 \times 10^{-4}$ and utilized a stepwise learning-rate schedule method that reduces the learning rate by a factor of 0.75 every 5 epochs based on our training observations and empirical evaluation, while for Agriculture-vison models, we use initial learning rates of $2.8 \times 10^{-4}$ and apply a cosine annealing scheduler that reduces the learning rates over epochs (for a maximum epoch of 40).

\subsection{Augmentation and evaluation methods}
During training, all data is sampled uniformly and augmented with random flip (horizontal and vertical), rotation ($90$ degree), Gaussian noise, and brightness contrast (all probabilities are 0.5) for each epoch. The albumentations library ~\citep{Albumentations2018} for data augmentation is utilized in this work. Please note that all training images are normalized to [0.0, 1.0] after data augmentation.

During test and evaluation, we apply test time augmentation (TTA) in terms of flipping and mirroring. For Vaihingen data, we use sliding windows (with $448 \times 448$ size at a 100-pixel stride) on a test image and stitch the results together by averaging the predictions of the over-lapping TTA regions to form the output. For the agriculture-vision data, we first apply TTA on the full size test image ($512 \times 512$) and average the predictions to get the final output. The performance is measured by the F1-score for Vaihingen dataset, and the modified Intersection over Union (IoU)~\citep{liu_00} for the agriculture-vision dataset. Please note that the mIoU metric was computed by averaging over the nine classes (including the 'Background' class) in the Agriculture-Vision benchmark dataset. 

\subsection{Test results} \label{exp-pots}
We tested our trained models on the hold out test sets of the Vaihingen and Agriculture-Vision datasets. The test results are shown in Table~\ref{tab:vaihingen_scores} and Table~\ref{tab:agri_scores}, respectively.  It is clearly visible for all the cases that our method outperforms all the state-of-the-art methods with a significant margin. For the Vaihingen dataset, it can be seen that the accuracy for the 'Car' class (90.8\% F1-score) is notably improved (+2.5\%) using our method in comparison to other methods. In case of the Agriculture-Vision dataset, many difficult classes also show significant increases in terms of IoU accuracies, e.g., double plant (+9.4\%), drydown (+5.8\%), endrow (+8.2\%), and planter skip (+5.3\%), etc. 

\begin{figure}[htpb!]
 \centering
  \includegraphics[width=1\textwidth]{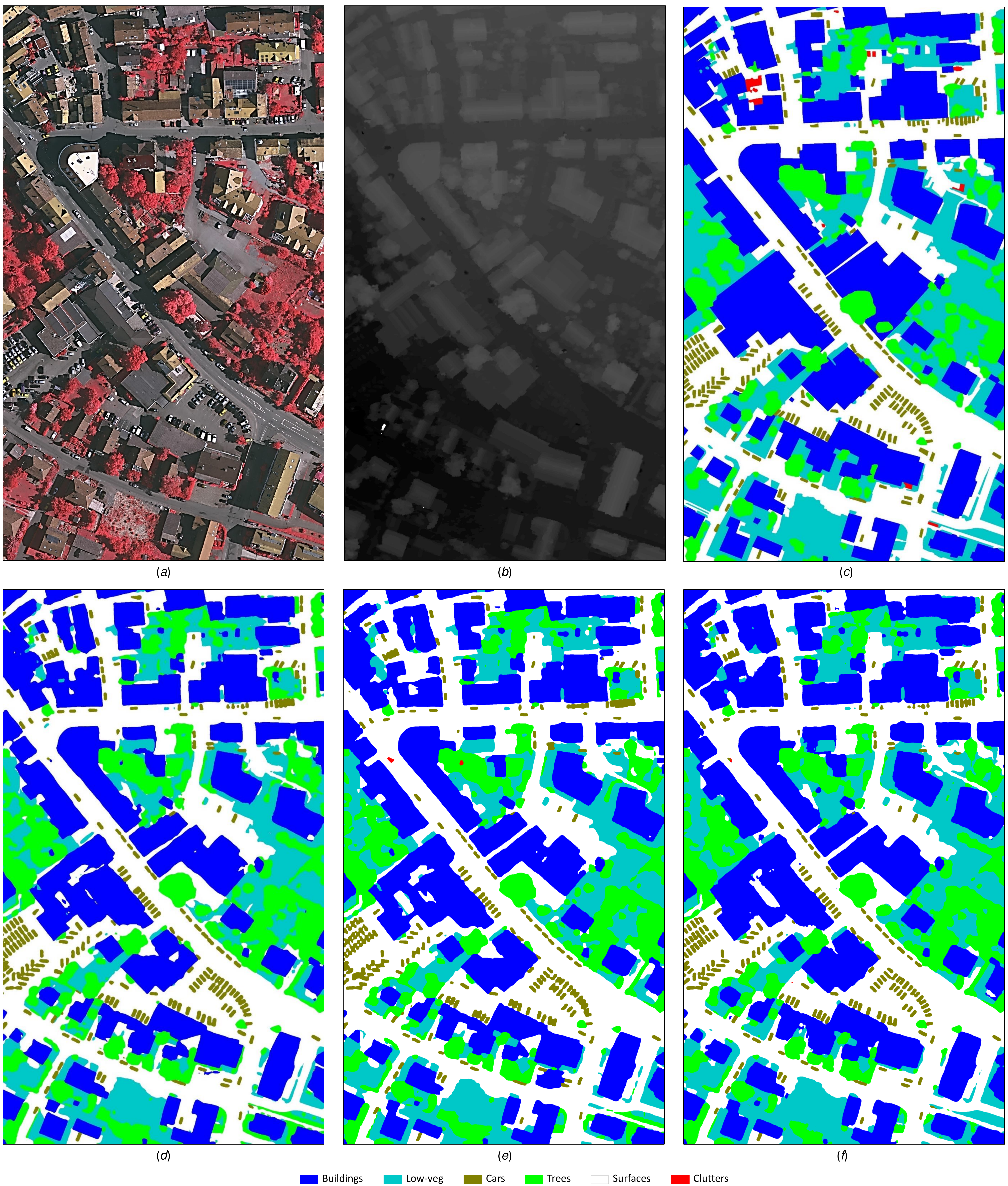}
  \caption{Segmentation results for the test image of Vaihingen tile-27: (a) the test IRRG image, (b) the DSM image, (c) the ground truth, (d) DDCM-R50, (e) SCG-GCN, (f) $\text{PAGNet}^{a}$}
  \label{fig:vaihingen_visual}
\end{figure}

\begin{figure}[htpb!]
 \centering
  \includegraphics[width=1\textwidth]{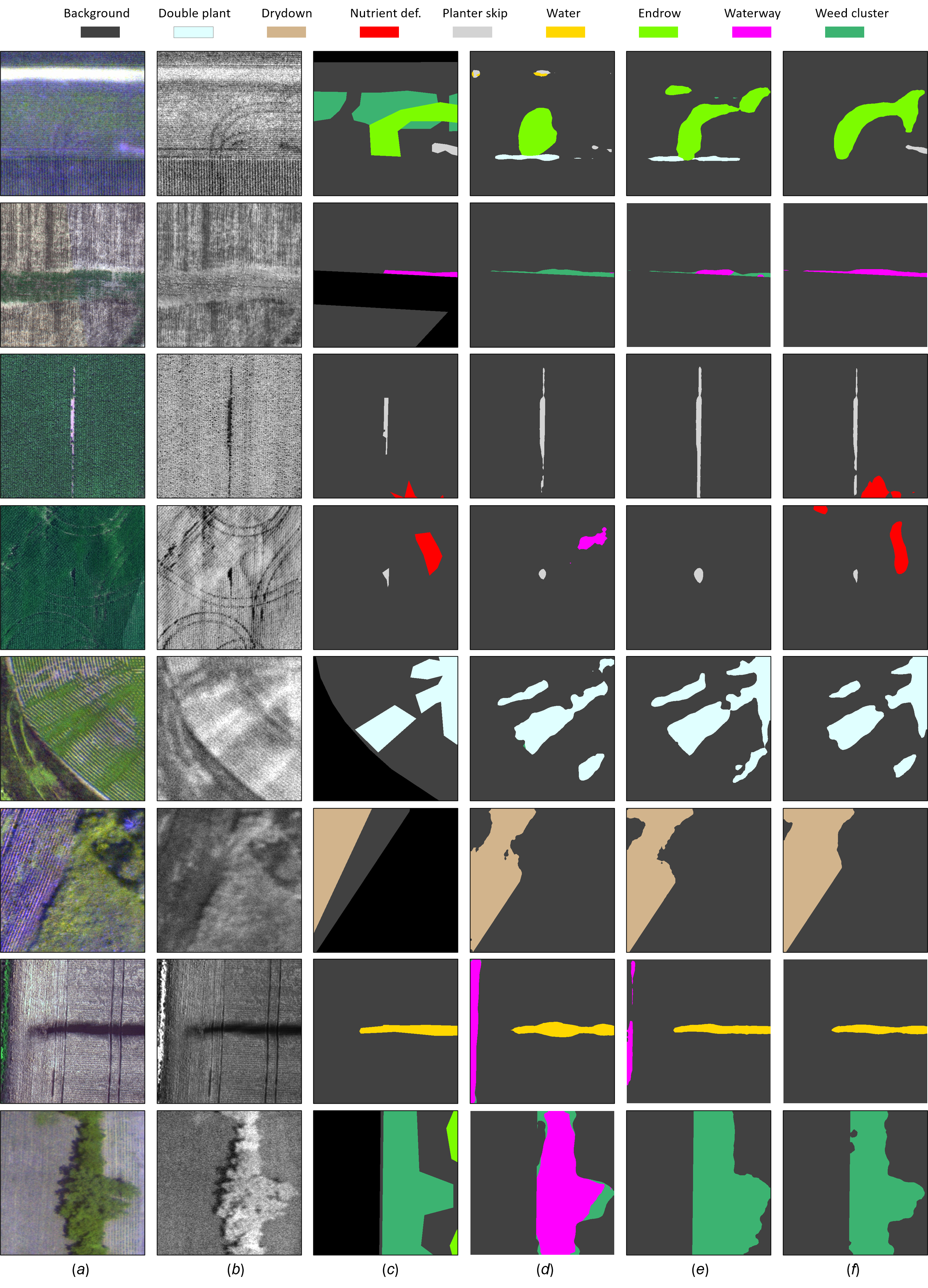}
  \caption{Segmentation results on the validation images of the agriculture-vision dataset: (a) the RGB images, (b) the NIR images, (c) the ground truth, (d) $\text{MSCG-Net50}$, (e) $\text{PAFNet}^{b}$, (f)  $\text{PAGNet}^{b}$}
  \label{fig:agri_visual}
\end{figure}

A qualitative comparison of the segmentation results from our trained models and the ground truths on the validation data are shown in Fig.~\ref{fig:vaihingen_visual} and Fig.~\ref{fig:agri_visual}. It can be visually verified that the classification maps obtained from our PAG-Net models tend to be less noisy and have smooth and fine-gained boundary recovery without any post-processing. In addition, our multi-modal $\text{PAGNet}^{b}$ model obtained the best performance on the Agriculture-Vision dataset with $48.2\%$ mIoU ($+4.2\%$) with fewer training parameters ($6.14$M) and $2\times$ faster training and inference speed on both CPU ($127$ms) and GPU ($11$ms) in comparison to MSCG-Net50 as shown in Table~\ref{tab:parameters}. It is worth noting that our two PAF-base unimodal models ($\text{PAFNet}^{a}$ and $\text{PAFNet}^{b}$) also obtain the best performance compared to other unimodal methods on both the Vaihingen and Agriculture-Vision datasets. 

\begin{table}[hptb!]
  \tbl{Comparisons between our method with other published methods on the hold-out test images of Vaihingen Dataset.}
{\resizebox{\columnwidth}{!}
{\begin{tabular}{lccccccc} \toprule
    \text{Models} & $\text{OA}$ & \text{Surface} & \text{Building} & \text{Low-veg} & \text{Tree} & \text{Car}  & \text{mF1} \\  \midrule 
    UOA~\citep{lin2016efficient}                 & 0.876  & 0.898  & 0.921  & 0.804 & 0.882  & 0.820  & 0.865 \\  
    DNN\_HCRF~\citep{liu2019semantic}            & 0.878  & 0.901  & 0.932  & 0.814 & 0.872  & 0.720  & 0.848\\ 
    ADL\_3~\citep{paisitkriangkrai2015effective} & 0.880  & 0.895  & 0.932  & 0.823 & 0.882  & 0.633  & 0.833 \\ 
    DST\_2~\citep{Sherrah16}                     & 0.891  & 0.905  & 0.937  & 0.834 & 0.892  & 0.726 & 0.859 \\ 
    ONE\_7~\citep{audebert2016semantic}          & 0.898  & 0.910  & 0.945  & {0.844} & 0.899  & 0.778 & 0.875\\ 
    DLR\_9~\citep{MarmanisSWGDS16}               & 0.903  & 0.924  & 0.952  & 0.839 & 0.899  & 0.812  & 0.885 \\  
    GSN~\citep{wang2017gated}                    & 0.903  & 0.922  & 0.951  & 0.837 & 0.899 & 0.824  & 0.887 \\ 
    RWSNet~\citep{Jiang2020RWSNetAS}             & 0.899  & 0.916  & 0.947  & 0.840 & 0.893 & 0.860  & 0.891 \\ 
    DDCM-R50~\citep{liu2020TGRS} & 0.904 & {0.927}& 0.953    & 0.833   &0.894 & 0.883 & 0.898  \\ 
     $\operatorname{SCG-GCN}$~\citep{Liu_2020_CVPR_Workshops} & 0.904  & 0.924  & 0.948  & \text{0.839} & \text{0.897}  & 0.880  & 0.898 \\ 
    FuseNet(IRRG+DSM/NDVI) & 0.908 & {0.913}& 0.943    & \textbf{0.848}  &0.899 & 0.859 & 0.901  \\  \hdashline
     $\text{PAFNet}^{a}$(DSM-IRRG) & 0.906  & 0.929 & 0.949  & 0.826 & 0.894  & 0.905  & 0.900 \\
     $\text{PAGNet}^{a}$(IRRG+DSM)  & \textbf{0.913}  & \textbf{0.930}  & \textbf{0.952}  & \text{0.843} & \textbf{0.900}  & \textbf{0.908}  & \textbf{0.907}  \\ 
     \bottomrule
\end{tabular}
} }
\label{tab:vaihingen_scores}%
\end{table}

\begin{table}[hptb!]
\tbl{Comparisons between our method with other published methods in terms of mIoUs and class IoUs on the hold-out Agriculture-Vision test set.}
{\resizebox{\textwidth}{!}{
\begin{tabular}{lcccccccccc} \toprule 
    \thead{Models} & \thead{mIoU} & \thead{Background} & \thead{Double \\plant} & \thead{Drydown} & \thead{Endrow} & \thead{Nutrient \\deficiency}  & \thead{Planter \\skip}  & \thead{Water} & \thead{Waterway}& \thead{Weed \\cluster} \\  \midrule 
     DeepLabv3(os=8) & 0.322 & 0.704  & 0.215 & 0.510 & 0.126  & 0.394  & 0.204 & 0.157  & 0.337 & 0.250 \\  
     DeepLabv3+(os=8) & 0.391 & 0.710  & 0.197 & 0.509 & 0.195  & 0.413  & 0.244 & 0.623  & 0.341 & 0.280 \\  
     DeepLabv3(os=16) & 0.422 & 0.727  & 0.252 & 0.536 & 0.210  & 0.440  & 0.246 & 0.704  & 0.386 & 0.299 \\  
     DeepLabv3+(os=16) & 0.424 & 0.725  & 0.260 & 0.536 & 0.241  & 0.442  & 0.244 & 0.703  & 0.379 & 0.288\\  
     FPN-ResNet~\citep{chiu2020agriculturevision}  & 0.437 & 0.726  & 0.279 & 0.523 & 0.243  & 0.438  & 0.310 & \textbf{0.713}  & \textbf{0.388} & 0.309\\  
     MSCG-Net50~\citep{Liu_2020_CVPR_Workshops} & 0.441 & 0.716  & 0.289 & 0.513 & 0.270  & 0.442  & 0.331 & 0.692  & 0.366 & 0.349\\  
    \hdashline 
     $\text{PAFNet}^{b}$(IR-RGB) & 0.442 & 0.687 & 0.343 & 0.562 & 0.281  & 0.420 & 0.305  & 0.680 & 0.378 & 0.324 \\ 
     $\text{PAGNet}^{b}$(RGB+IR) & \textbf{0.482}  & \textbf{0.740}  & \textbf{0.383} & \textbf{0.581}  & \textbf{0.352}  & \textbf{0.460}  & \textbf{0.384} & \text{0.686} & \text{0.373} & \textbf{0.379}\\ \hline
    \end{tabular}
} }
\label{tab:agri_scores}%
\end{table}

\newpage
\section{Discussion} \label{addtion}
In our network, pyramid attention fusion (PAF) modules are employed to capture multi-level and cross-view robust representations, and gated fusion units (GFU) are designed to bridge and interact among different modalities to better combine multi-modality information. To validate the effectiveness of these modules, we perform ablation experiments on the Vaihingen dataset.

\subsection{Effect of the pyramid attention fusion module}
In Table~\ref{tab:ablation_paf}, we evaluate our model's performance by removing the key components of the PAF module, i.e., pyramid cross-view encoder (PCE), attention construction and updating (ACU) module as shown in Fig.~\ref{fig:ae_encoder}. Since PCE and ACU are interdependent, we are not able to just use ACU without the PCE unit. We, therefore, evaluated the following two variations: 1) V-1: replacing both PCE and ACU with using multi-level concatenation fusion networks (FPN-style) with the same number of hidden features of the PAF module. 2) V-2: only removing ACU module. 
It is evidently shown that in absence of our pyramid attention fusion module, the V-1 model tends to underperform a lot on small objects (e.g., 'Car' $-4.8\%$), while only applying our pyramid cross-view encoder block, our V-2 model improves the performance a bit overall (mF1: $+0.9\%$ in contrast to V-1 model) but still underperforms on the small class 'Car' (mF1: $-2.4\%$ in contrast to PAGNet). 
\begin{table*}[hptb!]
  \tbl{The effect of the two key units (PCE, ACU) of our PAF module on the hold-out test images of Vaihingen Dataset.}
  {\resizebox{\columnwidth}{!}{
\begin{tabular}{c|cc|cc|cc|cc|cc|cc} \hline 
    Model  & PCE  & ACU & \text{OA} &  $\Delta\%$& \text{Building} & $\Delta\%$ & \text{Car} & $\Delta\%$ & \text{mF1} & $\Delta\%$ & Steps (K)\\  \hline   
     V-1  &  \xmark    & \xmark    & 0.898 & -1.5  &0.942  & -1.0 &0.860 & -4.8  & 0.890 & -1.7 & 31 \\ 
     V-2  &  \cmark   & \xmark    & 0.906 & -0.7  &0.945  & -0.7 &0.884 & -2.4 & 0.899 & -0.8 & 25 \\ \hline  
     PAGNet  &  \cmark    & \cmark   & \textbf{0.913} & -  & \textbf{0.952} &-  & \textbf{0.908} &-  & \textbf{0.907} & - & 29 \\\hline
\end{tabular}
} }
\label{tab:ablation_paf}%
\end{table*}

We also investigated the effect of the latent features and cross-view settings on the performance. Note that, we assume that the number of latent features ($c$) should be close to the number for classes (i.e., $6$ for Vaihingen dataset). The latent features are thus set to be in the range of $\{4, 6, 8, 12\}$, and the number of views ($v$) are in the range of $\{1, 2, 3\}$. Table~\ref{tab:ablation_kernel} presents the details of the evaluation results where five models are trained on various latent features and cross-view settings. Our model with latent features of 6 and view number of 3 achieves the best results. Note that the latent feature number does not show a significant impact on overall performance (mF1: $\pm 0.5\%$), while the number of views seems to be more sensitive on both overall results (mF1: $\pm 1.0\%$) and the performance on small objects (mF1: $\pm 2.9\%$).
\begin{table}[hptb!]
  \tbl{The effect of different the number of latent features and views of our PAF module on the hold-out Vaihingen test set.}
  {\resizebox{\columnwidth}{!}{
\begin{tabular}{c|cc|cc|cc|cc|cc|cc} \hline 
    \text{Model} & $c$  & $v$ & \text{OA} &  $\Delta\%$& \text{Building} & $\Delta\%$ & \text{Car} & $\Delta\%$ & \text{mF1} & $\Delta\%$ & Steps (K)\\  \hline     
     \text{PAG-v1} & 4    & 3    & 0.907 & -0.6  &0.950  & -0.2 &0.904 & -0.4  & 0.903 & -0.4 & 35 \\   
     \text{PAG-v2} & 8    & 3    & 0.910 & -0.3  &\textbf{0.957}  & +0.3 &0.905 & -0.3  & 0.904 & -0.3 & 21\\
     \text{PAG-v3} & 12   & 3    & 0.908 & -0.5  &0.952  &  0 &0.891 & -1.7  & 0.902  & -0.5 & 17 \\ 
     \text{PAG-v4} & 6    & 1    & 0.902 & -1.1  &0.947  & -0.5 &0.879 & -2.9  & 0.897 & -1.0 & 19 \\ 
     \text{PAG-v5} & 6    & 2    & 0.909 & -0.4  &0.949  & -0.3 &0.898 & -1.0  & 0.904 & -0.3 & 23\\ \hline  
     \text{PAGNet} & 6    & 3   & \textbf{0.913} & -  & \text{0.952} &-  & \textbf{0.908} &-  & \textbf{0.907} & - & 29 \\\hline
\end{tabular}
} }\tabnote{\textsuperscript{*}$c$ is the number of latent features and $v$ denotes the number of views. Here Steps K=$1000$ denote training iterations.}
\label{tab:ablation_kernel}%
\end{table}

\subsection{Effect of the gated fusion unit}
\begin{figure}[htpb!]
 \centering
  \includegraphics[width=0.8\textwidth]{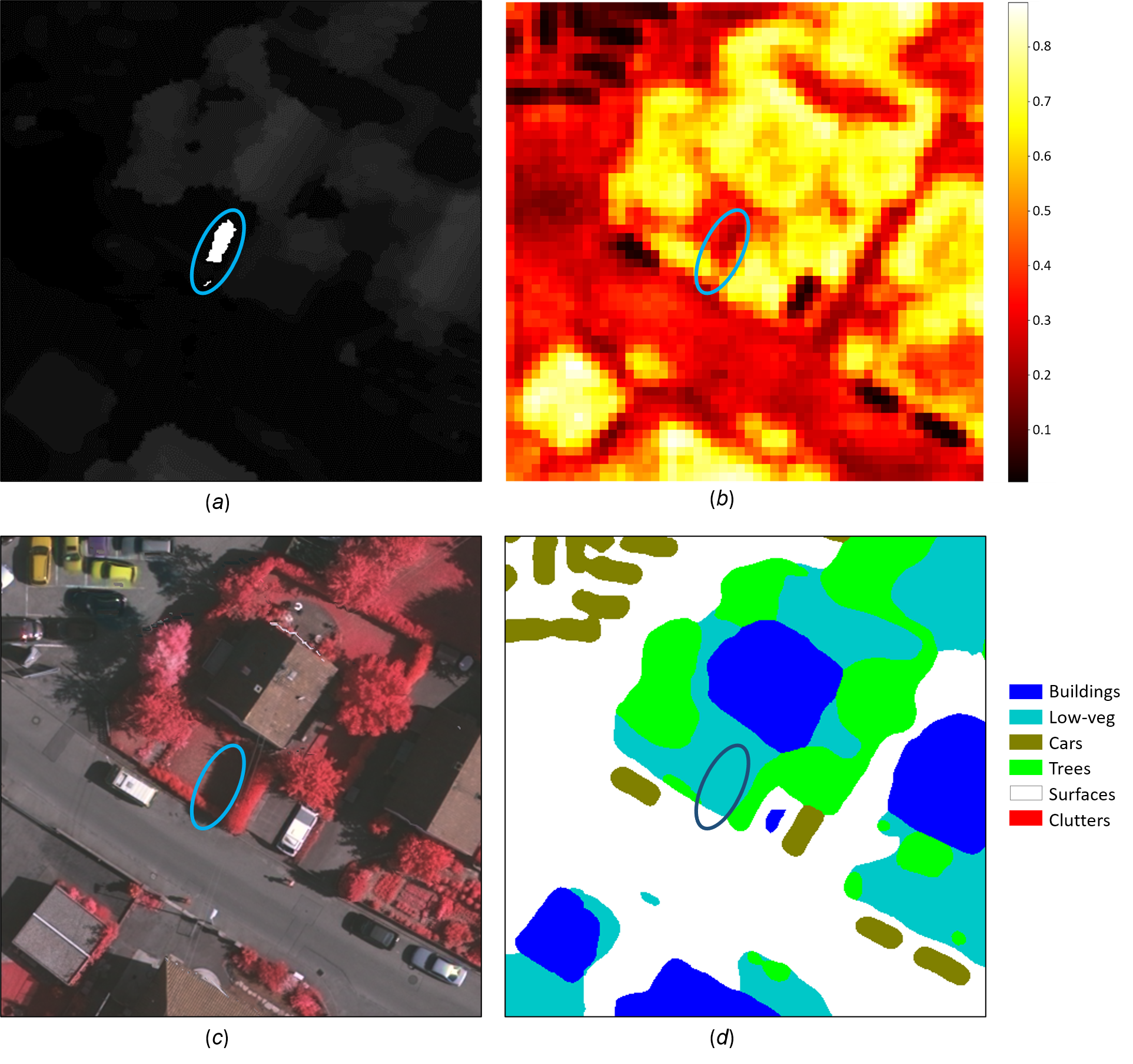}
  \caption{Heatmap of the GFU learned attention gate : (a) the DSM image (containing some noisy patch), (b) the attention gate heatmap, (c) the IRRG image, (d) the prediction.}
  \label{fig:heatmap}
\end{figure}

The GFU module plays a very important role for the effectiveness and efficiency of our multi-modal PAGNet. We, therefore, evaluate GFU by comparing it with the other two commonly used fusion methods (i.e., element-wise summing, and concatenation). 
Table~\ref{tab:ablation_gfu} displays the performance of these three methods. It is clearly shown that simply concatenating or summing multi-modality features will cause a degradation in performance to unimodal models. Our GFU approach, instead, shows notable performance gains (mF1: $+1.1\sim 1.4\%$) in general and significantly boosts the results on small objects (mF1: $+2.2\sim 3.3\%$) and improves the training converges speed (+$2\text{x}$ faster). We visualized the GFU module learned attention gate map as shown in Figure~\ref{fig:heatmap}. It illustrates that GFU module is able to capture a significant or complementary part of the information contained in the DSM data and diminish the influence of noisy data as well.
\begin{table*}[hptb!]
  \tbl{Test performance of different fusion settings on Vaihingen test set.}
{\resizebox{\columnwidth}{!}{
\begin{tabular}{ccc|cc|cc|cc|cc|cc} \hline
    {\textcircled{+}} & {\textcircled{c}} &{\textcircled{g}}& \text{OA} &  $\Delta\%$& \text{Building} & $\Delta\%$ & \text{Car} & $\Delta\%$ & \text{mF1} & $\Delta\%$ & Steps (K)& $\Delta$\\  \hline  
    \cmark  &  &   & 0.901 & -1.2  &0.950  & -0.2 &0.875  & -3.3  & 0.893 & -1.4 & 67 & 2.3x\\ 
     & \cmark &   & 0.906 & -0.7  &0.946  & -0.6 &0.886  &-2.2  & 0.896 & -1.1 & 92 & 3.1x\\ 
      &  & \cmark  & \textbf{0.913} & -  & \textbf{0.952} &-  & \textbf{0.908} &-  & \textbf{0.907} & - & \textbf{29} & -\\\hline
\end{tabular}
}}
\tabnote{\textsuperscript{*}{\textcircled{+}} denote point-wise summing fusion, {\textcircled{c}} is concatenation fusion, and {\textcircled{g}} is our gated fusion method.}
\label{tab:ablation_gfu}%
\end{table*}

\subsection{Effect of missing  and noisy data}
We also evaluate the performance of our model to handle situations where DSM data are missing, noisy and completely interfered during testing. Specifically, we assume that the IRRG data modality of the Vaihingen dataset is available, while the DSM modality is
missing (letting DSM data to 0), random noisy signal (using white noise data sampled from 0 to 255), or completely interfered (setting data value to 255). Table~\ref{tab:miss_noisy_data} illustrates the results. It clearly indicates that our model is capable of dealing with missing or noisy data. In other words, our model, which was trained with all data modalities (e.g., IRRG+DSM), generalizes well to circumstances in which extra modality data (e.g, DSM) is absent or entirely noisy during the testing phase. Our model's weakness for missing and noisy data modalities is that it does not handle missing primary modalities adequately (e.g., IRRG or RGB). In many situations, this is not an issue because it is usual to merely evaluate a small number of extra data modalities in remote sensing.

\begin{table}[hptb!]
  \tbl{Evaluation results with missing, noisy and interfered DSM data on the hold-out test images of Vaihingen Dataset.}
{\resizebox{\columnwidth}{!}
{\begin{tabular}{lccccccc} \toprule
    \text{Modalities} & $\text{OA}$ & \text{Surface} & \text{Building} & \text{Low-veg} & \text{Tree} & \text{Car}  & \text{mF1} \\  \midrule 
     $\text{Baseline (IRRG+DSM)}$  & \textbf{0.913}  & \textbf{0.930}  & \textbf{0.952}  & \textbf{0.843} & \textbf{0.900}  & \textbf{0.908}  & \textbf{0.907}  \\ \hdashline
     (IRRG+missing-DSM)  & \text{0.908}  & \text{0.926}  & \text{0.949}  & \text{0.839} & \text{0.897}  & \text{0.903}  & \text{0.903}  \\ 
     (IRRG+random-noisy-data)  & \text{0.904}  & \text{0.923}  & \text{0.943}  & \text{0.837} & \text{0.898}  & \text{0.900}  & \text{0.900}  \\ 
     (IRRG+interferred-data)  & \text{0.899}  & \text{0.905}  & \text{0.943}  & \text{0.836} & \text{0.892}  & \text{0.902}  & \text{0.896}  \\ 
     \bottomrule
\end{tabular}
} }
\label{tab:miss_noisy_data}%
\end{table}

\section{Conclusions} \label{concl}
We presented a novel pyramid attention and gated fusion method for multi-modality land cover and land use mapping in remote sensing. Our proposed pyramid attention fusion (PAF) module can effectively capture multi-level and cross-view attention maps to obtain rich and robust representations, that can further be flexibly harnessed as a key fusion bridge between multiple modalities using our developed gated fusion (GFU) algorithms. The GFU module can tune the noisy modalities and extract complementary features to improve the performance of our multimodal models. Built upon the PAF and GFU modules, our MultiModNet framework provides an end-to-end and lightweight multi-modal segmentation solution, which achieves the state-of-the-art performance and outperforms the strong baselines on two different representative remote sensing datasets. In addition, our methods easily generalize to more than two modalities for addressing more complicated problems in remote sensing.

\section*{Acknowledgements}
The benchmark datasets: the Vaihingen dataset was provided by the International Society for Photogrammetry and Remote Sensing (ISPRS); the Agriculture-Vision dataset was provided by UIUC, IntelinAir and University of Oregon. This work was supported by the foundation of the Research Council of Norway under Grant 272399 and Grant 309439. 

\section*{Disclosure statement}

No potential conflict of interest was reported by the authors.

\section*{Funding}

This work was supported by the foundation of Norges Forskningsråd under Grant 272399 and Grant 309439. 
 
\bibliographystyle{tfcad}
\bibliography{interactcadsample.bbl}

\begin{thebibliography}{54}
\newcommand{\enquote}[1]{``#1''}
\providecommand{\natexlab}[1]{#1}
\providecommand{\url}[1]{\normalfont{#1}}
\providecommand{\urlprefix}{}

\bibitem[Audebert, Le~Saux, and Lef{\`e}vre(2016)]{audebert2016semantic}
Audebert, Nicolas, Bertrand Le~Saux, and S{\'e}bastien Lef{\`e}vre. 2016.
  ``Semantic segmentation of earth observation data using multimodal and
  multi-scale deep networks.'' In \emph{Asian Conference on Computer Vision},
  180--196. Springer.

\bibitem[Audebert, Le~Saux, and Lef\`evre(2017)]{audebert2017joint}
Audebert, Nicolas, Bertrand Le~Saux, and S\'ebastien Lef\`evre. 2017. ``Joint
  learning from earth observation and {OpenStreetMap} data to get faster better
  semantic maps.'' In \emph{Proceedings of the IEEE Conference on Computer
  Vision and Pattern Recognition Workshops}, 67--75.

\bibitem[Audebert, Le~Saux, and Lef{\`e}vre(2018)]{audebert2018beyond}
Audebert, Nicolas, Bertrand Le~Saux, and S{\'e}bastien Lef{\`e}vre. 2018.
  ``Beyond RGB: Very high resolution urban remote sensing with multimodal deep
  networks.'' \emph{ISPRS J. Photogramm. Remote Sensing} 140: 20--32.

\bibitem[Audebert, Le~Saux, and Lef{\`e}vre(2019)]{audebert2019deep}
Audebert, Nicolas, Bertrand Le~Saux, and S{\'e}bastien Lef{\`e}vre. 2019.
  ``Deep learning for classification of hyperspectral data: A comparative
  review.'' \emph{IEEE Geoscience and Remote Sensing Magazine} 7 (2): 159--173.

\bibitem[Badrinarayanan, Kendall, and Cipolla(2017)]{badrinarayanan2017segnet}
Badrinarayanan, Vijay, Alex Kendall, and Roberto Cipolla. 2017. ``SegNet: A
  Deep Convolutional Encoder-Decoder Architecture for Image Segmentation.''
  \emph{IEEE Transactions on Pattern Analysis and Machine Intelligence} 39
  (12): 2481--2495.

\bibitem[Bello and Aina(2014)]{bello2014satellite}
Bello, Olalekan~Mumin, and Yusuf~Adedoyin Aina. 2014. ``Satellite remote
  sensing as a tool in disaster management and sustainable development: towards
  a synergistic approach.'' \emph{Procedia-Social and Behavioral Sciences} 120:
  365--373.

\bibitem[Buslaev et~al.(2020)]{Albumentations2018}
Buslaev, Alexander, Vladimir~I. Iglovikov, Eugene Khvedchenya, Alex Parinov,
  Mikhail Druzhinin, and Alexandr~A. Kalinin. 2020. ``Albumentations: Fast and
  Flexible Image Augmentations.'' \emph{Information} 11 (2).
  \urlprefix\url{https://www.mdpi.com/2078-2489/11/2/125}.

\bibitem[Chen et~al.(2018)]{chen2018deeplab}
Chen, Liang-Chieh, George Papandreou, Iasonas Kokkinos, Kevin Murphy, and
  Alan~L Yuille. 2018. ``{DeepLab}: Semantic Image Segmentation with Deep
  Convolutional Nets, Atrous Convolution, and Fully Connected CRFs.''
  \emph{IEEE Trans. Pattern Anal. Mach. Intell.} 40 (4): 834--848.

\bibitem[Chiu et~al.(2020{\natexlab{a}})]{liu_00}
Chiu, Mang~Tik, Xingqian Xu, Kai Wang, Jennifer Hobbs, Naira Hovakimyan,
  Thomas~S. Huang, Honghui Shi, et~al. 2020{\natexlab{a}}. ``The 1st
  Agriculture-Vision Challenge: Methods and Results.'' In \emph{the IEEE/CVF
  Conference on Computer Vision and Pattern Recognition Workshops (CVPRW)},
  212--218.

\bibitem[Chiu et~al.(2020{\natexlab{b}})]{chiu2020agriculturevision}
Chiu, Mang~Tik, Xingqian Xu, Yunchao Wei, Zilong Huang, Alexander~G. Schwing,
  Robert Brunner, Hrant Khachatrian, et~al. 2020{\natexlab{b}}.
  ``Agriculture-Vision: A Large Aerial Image Database for Agricultural Pattern
  Analysis.'' In \emph{the IEEE/CVF Conference on Computer Vision and Pattern
  Recognition (CVPR)}, June.

\bibitem[Couprie et~al.(2013)]{couprie2013indoor}
Couprie, Camille, Cl{\'e}ment Farabet, Laurent Najman, and Yann LeCun. 2013.
  ``Indoor semantic segmentation using depth information.'' \emph{arXiv
  preprint arXiv:1301.3572} .

\bibitem[Dosovitskiy et~al.(2021)]{dosovitskiy2020vit}
Dosovitskiy, Alexey, Lucas Beyer, Alexander Kolesnikov, Dirk Weissenborn,
  Xiaohua Zhai, Thomas Unterthiner, Mostafa Dehghani, et~al. 2021. ``An Image
  is Worth 16x16 Words: Transformers for Image Recognition at Scale.''
  \emph{ICLR} .

\bibitem[Fan et~al.(2021)]{fan2021disaster}
Fan, Chao, Cheng Zhang, Alex Yahja, and Ali Mostafavi. 2021. ``Disaster City
  Digital Twin: A vision for integrating artificial and human intelligence for
  disaster management.'' \emph{International Journal of Information Management}
  56: 102049.

\bibitem[Feng et~al.(2019)]{feng2019multisource}
Feng, Quanlong, Dehai Zhu, Jianyu Yang, and Baoguo Li. 2019. ``Multisource
  hyperspectral and lidar data fusion for urban land-use mapping based on a
  modified two-branch convolutional neural network.'' \emph{ISPRS International
  Journal of Geo-Information} 8 (1): 28.

\bibitem[{Fu} et~al.(2019)]{DANet}
{Fu}, J., J.~{Liu}, H.~{Tian}, Y.~{Li}, Y.~{Bao}, Z.~{Fang}, and H.~{Lu}. 2019.
  ``Dual Attention Network for Scene Segmentation.'' In \emph{2019 IEEE/CVF
  Conference on Computer Vision and Pattern Recognition (CVPR)}, 3141--3149.

\bibitem[Fu et~al.(2019)]{fu2019dual}
Fu, Jun, Jing Liu, Haijie Tian, Yong Li, Yongjun Bao, Zhiwei Fang, and Hanqing
  Lu. 2019. ``Dual attention network for scene segmentation.'' In
  \emph{Proceedings of the IEEE/CVF Conference on Computer Vision and Pattern
  Recognition}, 3146--3154.

\bibitem[Ghosh et~al.(2018)]{ghosh2018stacked}
Ghosh, Arthita, Max Ehrlich, Sohil Shah, Larry Davis, and Rama Chellappa. 2018.
  ``{Stacked U-Nets for Ground Material Segmentation in Remote Sensing
  Imagery}.'' In \emph{2018 IEEE/CVF Conference on Computer Vision and Pattern
  Recognition Workshops (CVPRW)}, 252--2524.

\bibitem[G{\'o}mez-Chova et~al.(2015)]{gomez2015multimodal}
G{\'o}mez-Chova, Luis, Devis Tuia, Gabriele Moser, and Gustau Camps-Valls.
  2015. ``Multimodal classification of remote sensing images: A review and
  future directions.'' \emph{Proceedings of the IEEE} 103 (9): 1560--1584.

\bibitem[Hazirbas et~al.(2016)]{hazirbas2016fusenet}
Hazirbas, Caner, Lingni Ma, Csaba Domokos, and Daniel Cremers. 2016. ``Fusenet:
  Incorporating depth into semantic segmentation via fusion-based cnn
  architecture.'' In \emph{Asian Conference on Computer Vision}, 213--228.
  Springer.

\bibitem[Hong et~al.(2020)]{hong2020more}
Hong, Danfeng, Lianru Gao, Naoto Yokoya, Jing Yao, Jocelyn Chanussot, Qian Du,
  and Bing Zhang. 2020. ``More diverse means better: Multimodal deep learning
  meets remote-sensing imagery classification.'' \emph{IEEE Transactions on
  Geoscience and Remote Sensing} .

\bibitem[Howard et~al.(2019)]{mobilev3}
Howard, Andrew, Mark Sandler, Grace Chu, Liang-Chieh Chen, Bo~Chen, Mingxing
  Tan, Weijun Wang, et~al. 2019. ``Searching for mobilenetv3.'' In
  \emph{Proceedings of the IEEE/CVF International Conference on Computer
  Vision}, 1314--1324.

\bibitem[Hu, Shen, and Sun(2018)]{hu2018squeeze}
Hu, Jie, Li~Shen, and Gang Sun. 2018. ``Squeeze-and-excitation networks.'' In
  \emph{Proceedings of the IEEE Conference on Computer Vision and Pattern
  Recognition}, 7132--7141.

\bibitem[Jiang et~al.(2020)]{Jiang2020RWSNetAS}
Jiang, Jie, Chengjin Lyu, Siying Liu, Y.~He, and Xuetao Hao. 2020. ``RWSNet: a
  semantic segmentation network based on SegNet combined with random walk for
  remote sensing.'' \emph{International Journal of Remote Sensing} 41: 487 --
  505.

\bibitem[Kampffmeyer, Salberg, and Jenssen(2016)]{kampffmeyer2016semantic}
Kampffmeyer, Michael, Arnt-Borre Salberg, and Robert Jenssen. 2016. ``Semantic
  segmentation of small objects and modeling of uncertainty in urban remote
  sensing images using deep convolutional neural networks.'' In
  \emph{Proceedings of the IEEE Conference on Computer Vision and Pattern
  Recognition Workshops}, 1--9.

\bibitem[Kampffmeyer, Salberg, and Jenssen(2018)]{michael2018}
Kampffmeyer, Michael, Arnt-B{\o}rre Salberg, and Robert Jenssen. 2018. ``Urban
  Land Cover Classification With Missing Data Modalities Using Deep
  Convolutional Neural Networks.'' \emph{IEEE Journal of Selected Topics in
  Applied Earth Observations and Remote Sensing} 11 (6): 1758--1768.

\bibitem[Kingma and Ba(2014)]{KingmaB14adam}
Kingma, Diederik~P., and Jimmy Ba. 2014. ``Adam: {A} Method for Stochastic
  Optimization.'' \emph{CoRR} abs/1412.6980.
  \urlprefix\url{http://arxiv.org/abs/1412.6980}.

\bibitem[Kipf and Welling(2016)]{kipf2016semi}
Kipf, Thomas~N, and Max Welling. 2016. ``Semi-supervised classification with
  graph convolutional networks.'' \emph{arXiv preprint arXiv:1609.02907} .

\bibitem[Li et~al.(2020)]{li2020multimodal}
Li, Xiao, Lin Lei, Yuli Sun, Ming Li, and Gangyao Kuang. 2020. ``Multimodal
  bilinear fusion network with second-order attention-based channel selection
  for land cover classification.'' \emph{IEEE Journal of Selected Topics in
  Applied Earth Observations and Remote Sensing} 13: 1011--1026.

\bibitem[Lin et~al.(2016)]{lin2016efficient}
Lin, Guosheng, Chunhua Shen, Anton Van Den~Hengel, and Ian Reid. 2016.
  ``Efficient piecewise training of deep structured models for semantic
  segmentation.'' In \emph{Proceedings of the IEEE Conference on Computer
  Vision and Pattern Recognition}, 3194--3203.

\bibitem[Lin et~al.(2017)]{lin2017feature}
Lin, Tsung-Yi, Piotr Doll{\'a}r, Ross Girshick, Kaiming He, Bharath Hariharan,
  and Serge Belongie. 2017. ``Feature pyramid networks for object detection.''
  In \emph{Proceedings of the IEEE Conference on Computer Vision and Pattern
  Recognition}, 2117--2125.

\bibitem[{Liu} et~al.(2020)]{liu2020TGRS}
{Liu}, Q., M.~{Kampffmeyer}, R.~{Jenssen}, and A.~B. {Salberg}. 2020. ``Dense
  Dilated Convolutions’ Merging Network for Land Cover Classification.''
  \emph{IEEE Transactions on Geoscience and Remote Sensing} 58 (9): 6309--6320.

\bibitem[Liu et~al.(2020{\natexlab{a}})]{liu2020scg}
Liu, Qinghui, Michael Kampffmeyer, Robert Jenssen, and Arnt-Børre Salberg.
  2020{\natexlab{a}}. ``{Self-Constructing Graph Convolutional Networks for
  Semantic Labeling}.'' In \emph{Proceedings of IGARSS 2020 - 2020 IEEE
  International Geoscience and Remote Sensing Symposium}, .

\bibitem[Liu et~al.(2020{\natexlab{b}})]{Liu_2020_CVPR_Workshops}
Liu, Qinghui, Michael~C. Kampffmeyer, Robert Jenssen, and Arnt-Børre Salberg.
  2020{\natexlab{b}}. ``{Multi-View Self-Constructing Graph Convolutional
  Networks With Adaptive Class Weighting Loss for Semantic Segmentation}.'' In
  \emph{the IEEE/CVF Conference on Computer Vision and Pattern Recognition
  (CVPR) Workshops}, June.

\bibitem[Liu et~al.(2019)]{liu2019semantic}
Liu, Yansong, Sankaranarayanan Piramanayagam, Sildomar~T Monteiro, and Eli
  Saber. 2019. ``Semantic segmentation of multisensor remote sensing imagery
  with deep ConvNets and higher-order conditional random fields.''
  \emph{Journal of Applied Remote Sensing} 13 (1): 016501.

\bibitem[Long, Shelhamer, and Darrell(2015)]{long2015fully}
Long, Jonathan, Evan Shelhamer, and Trevor Darrell. 2015. ``Fully Convolutional
  Networks for Semantic Segmentation.'' In \emph{Proceedings of the IEEE
  Conference on Computer Vision and Pattern Recognition}, 3431--3440.

\bibitem[Maggiori et~al.(2017)]{maggiori2017can}
Maggiori, Emmanuel, Yuliya Tarabalka, Guillaume Charpiat, and Pierre Alliez.
  2017. ``Can semantic labeling methods generalize to any city? the inria
  aerial image labeling benchmark.'' In \emph{2017 IEEE International
  Geoscience and Remote Sensing Symposium (IGARSS)}, 3226--3229. IEEE.

\bibitem[Marmanis et~al.(2016)]{MarmanisSWGDS16}
Marmanis, Dimitrios, Konrad Schindler, Jan~Dirk Wegner, Silvano Galliani, Mihai
  Datcu, and Uwe Stilla. 2016. ``Classification With an Edge: Improving
  Semantic Image Segmentation with Boundary Detection.'' \emph{CoRR}
  abs/1612.01337.  \urlprefix\url{http://arxiv.org/abs/1612.01337}.

\bibitem[Mohla et~al.(2020)]{FusAtNet}
Mohla, Satyam, Shivam Pande, Biplab Banerjee, and Subhasis Chaudhuri. 2020.
  ``FusAtNet: Dual Attention based SpectroSpatial Multimodal Fusion Network for
  Hyperspectral and LiDAR Classification.'' In \emph{2020 IEEE/CVF Conference
  on Computer Vision and Pattern Recognition Workshops (CVPRW)}, 416--425.

\bibitem[Mou and Zhu(2019)]{mou2019learning}
Mou, Lichao, and Xiao~Xiang Zhu. 2019. ``Learning to pay attention on spectral
  domain: A spectral attention module-based convolutional network for
  hyperspectral image classification.'' \emph{IEEE Transactions on Geoscience
  and Remote Sensing} 58 (1): 110--122.

\bibitem[Noor, Abdullah, and Hashim(2018)]{noor2018remote}
Noor, Norzailawati~Mohd, Alias Abdullah, and Mazlan Hashim. 2018. ``Remote
  sensing UAV/drones and its applications for urban areas: a review.'' In
  \emph{IOP conference series: Earth and environmental science}, Vol. 169,
  012003. IOP Publishing.

\bibitem[Paisitkriangkrai et~al.(2015)]{paisitkriangkrai2015effective}
Paisitkriangkrai, Sakrapee, Jamie Sherrah, Pranam Janney, Van-Den Hengel,
  et~al. 2015. ``Effective semantic pixel labelling with convolutional networks
  and conditional random fields.'' In \emph{Proceedings of the IEEE Conference
  on Computer Vision and Pattern Recognition Workshops}, 36--43.

\bibitem[Pashaei et~al.(2020)]{pashaei2020review}
Pashaei, Mohammad, Hamid Kamangir, Michael~J Starek, and Philippe Tissot. 2020.
  ``Review and evaluation of deep learning architectures for efficient land
  cover mapping with UAS hyper-spatial imagery: A case study over a wetland.''
  \emph{Remote Sensing} 12 (6): 959.

\bibitem[Ronneberger, Fischer, and Brox(2015)]{ronneberger2015u}
Ronneberger, Olaf, Philipp Fischer, and Thomas Brox. 2015. ``{U-Net}:
  Convolutional networks for biomedical image segmentation.'' In
  \emph{International Conference on Medical Image Computing and
  Computer-Assisted Intervention}, 234--241. Springer.

\bibitem[Rottensteiner et~al.(2012)]{isprs2012}
Rottensteiner, Franz, Gunho Sohn, Jaewook Jung, Markus Gerke, Caroline
  Baillard, Sebastien Benitez, and Uwe Breitkopf. 2012. ``The ISPRS benchmark
  on urban object classification and 3D building reconstruction.'' \emph{ISPRS
  Annals of the Photogrammetry, Remote Sensing and Spatial Information Sciences
  I-3 (2012), Nr. 1} 1 (1): 293--298.

\bibitem[Salberg(2011)]{ab_00}
Salberg, Arnt-Børre. 2011. ``Land Cover Classification of Cloud-Contaminated
  Multitemporal High-Resolution Images.'' \emph{IEEE Transactions on Geoscience
  and Remote Sensing} 49 (1): 377--387.

\bibitem[Salberg, Rudjord, and Solberg(2014)]{ab_01}
Salberg, Arnt-Børre, Øystein Rudjord, and Anne H.~Schistad Solberg. 2014.
  ``Oil Spill Detection in Hybrid-Polarimetric SAR Images.'' \emph{IEEE
  Transactions on Geoscience and Remote Sensing} 52 (10): 6521--6533.

\bibitem[Sherrah(2016)]{Sherrah16}
Sherrah, Jamie. 2016. ``Fully Convolutional Networks for Dense Semantic
  Labelling of High-Resolution Aerial Imagery.'' \emph{CoRR} abs/1606.02585.
  \urlprefix\url{http://arxiv.org/abs/1606.02585}.

\bibitem[Vaswani et~al.(2017)]{Attentionisall}
Vaswani, Ashish, Noam Shazeer, Niki Parmar, Jakob Uszkoreit, Llion Jones,
  Aidan~N Gomez, \L~ukasz Kaiser, and Illia Polosukhin. 2017. ``Attention is
  All you Need.'' In \emph{Advances in Neural Information Processing Systems
  30},  edited by I.~Guyon, U.~V. Luxburg, S.~Bengio, H.~Wallach, R.~Fergus,
  S.~Vishwanathan, and R.~Garnett, 5998--6008. Curran Associates, Inc.
  \urlprefix\url{http://papers.nips.cc/paper/7181-attention-is-all-you-need.pdf}.

\bibitem[Wang et~al.(2017)]{wang2017gated}
Wang, Hongzhen, Ying Wang, Qian Zhang, Shiming Xiang, and Chunhong Pan. 2017.
  ``Gated convolutional neural network for semantic segmentation in
  high-resolution images.'' \emph{Remote Sensing} 9 (5): 446.

\bibitem[Xu et~al.(2017)]{xu2017multisource}
Xu, Xiaodong, Wei Li, Qiong Ran, Qian Du, Lianru Gao, and Bing Zhang. 2017.
  ``Multisource remote sensing data classification based on convolutional
  neural network.'' \emph{IEEE Transactions on Geoscience and Remote Sensing}
  56 (2): 937--949.

\bibitem[Xu, Du, and Zhang(2018)]{fusion_xu}
Xu, Yonghao, Bo~Du, and Liangpei Zhang. 2018. ``Multi-Source Remote Sensing
  Data Classification via Fully Convolutional Networks and Post-Classification
  Processing.'' In \emph{IGARSS 2018 - 2018 IEEE International Geoscience and
  Remote Sensing Symposium}, 3852--3855.

\bibitem[Xu et~al.(2019)]{fusion_contest}
Xu, Yonghao, Bo~Du, Liangpei Zhang, Daniele Cerra, Miguel Pato, Emiliano
  Carmona, Saurabh Prasad, Naoto Yokoya, Ronny Hänsch, and Bertrand Le~Saux.
  2019. ``Advanced Multi-Sensor Optical Remote Sensing for Urban Land Use and
  Land Cover Classification: Outcome of the 2018 IEEE GRSS Data Fusion
  Contest.'' \emph{IEEE Journal of Selected Topics in Applied Earth
  Observations and Remote Sensing} 12 (6): 1709--1724.

\bibitem[Yuan, Chen, and Wang(2020)]{YuanCW19}
Yuan, Yuhui, Xilin Chen, and Jingdong Wang. 2020. ``Object-Contextual
  Representations for Semantic Segmentation.'' In \emph{Proceedings of the
  European Conference on Computer Vision (ECCV)}, 173--190.

\bibitem[Zhao et~al.(2017)]{zhao2016pyramid}
Zhao, Hengshuang, Jianping Shi, Xiaojuan Qi, Xiaogang Wang, and Jiaya Jia.
  2017. ``Pyramid Scene Parsing Network.'' In \emph{2017 IEEE Conference on
  Computer Vision and Pattern Recognition (CVPR)}, 6230--6239.

\end{thebibliography}

\end{document}